\documentclass[11pt]{article}

\usepackage[margin=1in]{geometry}
\usepackage{graphicx}
\usepackage{booktabs}
\usepackage{amsmath}
\usepackage[hidelinks]{hyperref}
\usepackage[round]{natbib}
\usepackage{caption}
\usepackage{tikz}
\usetikzlibrary{positioning,arrows.meta,fit,backgrounds,calc}
\usepackage{fontawesome5}
\usepackage{parskip}
\usepackage{microtype}
\definecolor{ink}{RGB}{45,45,45}
\definecolor{nbk}{RGB}{186,92,18}
\definecolor{prov}{RGB}{24,92,158}
\colorlet{paper}{black!2}
\definecolor{sq}{HTML}{197FB9}
\definecolor{re}{HTML}{D96D19}
\newcommand{\figttl}{\footnotesize\bfseries}
\title{K-Dense BYOK: An Open-Source AI Research Assistant That Runs Locally and Keeps a Hash-Chained Lab Notebook}

\author{Aubrey M. Brueckner, Darshil Patel, Yuhuan He, and Timothy Kassis\textsuperscript{*}\\
K-Dense Inc.\\
\textsuperscript{*}Correspondence: \texttt{timothy.kassis@k-dense.ai}}

\date{}

\begin{document}
\maketitle

\begin{center}
\begin{tikzpicture}[
  x=1mm, y=1mm, font=\sffamily\scriptsize, execute at begin node={\hyphenpenalty=10000\relax},
  every node/.style={text=ink, inner sep=0pt},
  panel/.style={draw=black!18, fill=black!3, rounded corners=2.2pt, line width=0.4pt},
  ptitle/.style={anchor=north west, font=\sffamily\footnotesize\bfseries, align=left},
  psub/.style={anchor=north west, font=\sffamily\tiny, text=black!60, align=left},
  badge/.style={circle, fill=ink, text=white, font=\sffamily\tiny\bfseries, minimum size=3.8mm,
                inner sep=0pt},
  chip/.style={draw=black!30, fill=white, rounded corners=1.2pt, inner sep=1.6pt, align=center,
               line width=0.35pt},
  body/.style={anchor=north west, align=left, text=black!70, font=\sffamily\tiny},
  card/.style={rounded corners=1.8pt, line width=0.6pt, inner sep=0pt},
  nextarrow/.style={-{Triangle[length=2.4mm,width=3mm]}, line width=1.1pt, draw=black!40},
]
\node[anchor=north west, font=\sffamily\large\bfseries] at (0.15,0)
  {Bring your own model. Keep a record you can check.};
\node[anchor=north east, chip, font=\sffamily\tiny, text=black!65, inner sep=2pt] at (164.85,-0.8)
  {\faUnlock\ \ Free \ · \ open source \ · \ MIT license};

\draw[panel] (0.15,-8) rectangle (46.5,-63.5);
\draw[panel] (51,-8) rectangle (110,-63.5);
\draw[panel] (115,-8) rectangle (164.85,-63.5);
\draw[nextarrow] (47,-35.75) -- (50.5,-35.75);
\draw[nextarrow] (110.6,-35.75) -- (114.4,-35.75);

\node[badge] at (4,-11.8) {1};
\node[ptitle] at (7.5,-9.9) {Your model, your machine};
\node[font=\Huge, text=black!55] at (23.25,-21.8) {\faLaptop};
\node[chip, anchor=north, text width=40mm, font=\sffamily\tiny, align=left] at (23.25,-30.2)
  {{\scriptsize\color{black!55}\faCloud}\ \ a hosted model, reached through your own account};
\node[chip, anchor=north, text width=40mm, font=\sffamily\tiny, align=left] at (23.25,-37.3)
  {{\scriptsize\color{black!55}\faDesktop}\ \ or a model on your own machine: nothing leaves it};
\node[body, text width=42.5mm] at (2,-45.6)
  {The model is chosen for each chat, so a sensitive analysis can run locally while a
   literature search uses a hosted model.};
\node[body, text width=42.5mm] at (2,-54.2)
  {One project is one ordinary folder. We host nothing, and the folder can be read years later
   without the app.};

\node[badge] at (55,-11.8) {2};
\node[ptitle] at (58.5,-9.9) {Kady does the work};
\node[psub] at (58.5,-13.8) {with scientific scaffolding, writing two records};
\foreach \cx/\num/\lab in {59.2/163/skills, 73.4/326/workflows, 87.6/229/data\\resources,
                           101.8/21/specialists} {
  \node[chip, minimum width=13.2mm, minimum height=8.6mm, font=\sffamily\tiny, text width=12mm]
    at (\cx,-22.2) {{\small\bfseries \num}\\[-0.2ex]\lab};
}
\draw[card, draw=nbk, fill=nbk!6] (52.7,-29.5) rectangle (80.3,-61.5);
\node[anchor=north east, font=\normalsize, text=nbk] at (78.9,-30.8) {\faBook};
\node[anchor=north west, font=\sffamily\tiny\bfseries, text=nbk] at (54.3,-31.2)
  {Living Lab Notebook};
\node[anchor=north west, font=\sffamily\tiny\itshape, text=black!55] at (54.3,-34.6)
  {what Kady reasoned};
\node[body, text width=24mm] at (54.3,-39.6)
  {Hypotheses, methods, observations and decisions, linked into an argument.};
\node[body, text width=24mm, font=\sffamily\tiny\bfseries, text=ink] at (54.3,-54)
  {Added to, never erased.};
\draw[card, draw=prov, fill=prov!6] (81.1,-29.5) rectangle (108.7,-61.5);
\node[anchor=north east, font=\normalsize, text=prov] at (107.3,-30.8) {\faFingerprint};
\node[anchor=north west, font=\sffamily\tiny\bfseries, text=prov] at (82.7,-31.2)
  {Provenance log};
\node[anchor=north west, font=\sffamily\tiny\itshape, text=black!55] at (82.7,-34.6)
  {what actually happened};
\node[body, text width=24mm] at (82.7,-39.6)
  {The app watches every action and fingerprints every file it touched.};
\node[body, text width=24mm, font=\sffamily\tiny\bfseries, text=ink] at (82.7,-54)
  {\textcolor{prov}{\faLock}\ Kady has no tool that can write here.};

\node[badge] at (119,-11.8) {3};
\node[ptitle] at (122.5,-9.9) {Tested on 20 prompts};
\node[psub, text width=46mm] at (117,-14.4)
  {vs.\ two managed platforms, one rubric fixed in advance, scored by the authors; mean \% of
   available points};
\fill[sq] (117,-24.2) rectangle (119.2,-22.8);
\node[anchor=west, font=\sffamily\tiny, text=black!70] at (119.9,-23.5) {Scientific quality};
\fill[re] (141.5,-24.2) rectangle (143.7,-22.8);
\node[anchor=west, font=\sffamily\tiny, text=black!70] at (144.2,-23.5) {Research execution};
\def\base{-42.8}
\foreach \cx/\q/\e/\name in {124/86.0/64.5/K-Dense BYOK, 140.5/74.8/26.0/Claude Science,
                             157/64.9/38.3/Biomni Lab} {
  \pgfmathsetmacro{\hq}{\q*0.15}
  \pgfmathsetmacro{\he}{\e*0.15}
  \fill[sq] (\cx-4.7,\base) [rounded corners=0.5mm] -- (\cx-4.7,\base+\hq) -- (\cx-0.4,\base+\hq)
            [sharp corners] -- (\cx-0.4,\base) -- cycle;
  \fill[re] (\cx+0.4,\base) [rounded corners=0.5mm] -- (\cx+0.4,\base+\he) -- (\cx+4.7,\base+\he)
            [sharp corners] -- (\cx+4.7,\base) -- cycle;
  \node[anchor=south, font=\sffamily\tiny, text=black!70] at (\cx-2.55,\base+\hq+0.4) {\q};
  \node[anchor=south, font=\sffamily\tiny, text=black!70] at (\cx+2.55,\base+\he+0.4) {\e};
  \node[anchor=north, font=\sffamily\tiny, align=center, text=black!70, text width=15mm]
    at (\cx,\base-0.8) {\name};
}
\draw[black!35, line width=0.3pt] (116.5,\base) -- (163.5,\base);
\node[chip, anchor=north west, align=left, text width=21.3mm, inner sep=2pt] at (117,-49)
  {{\normalsize\bfseries\color{prov} 20\,/\,20}\\[0.1ex]{\tiny bundles recorded the software environment}};
\node[chip, anchor=north west, align=left, text width=21.3mm, inner sep=2pt] at (141.2,-49)
  {{\normalsize\bfseries\color{black!55} 0\,/\,20}\\[0.1ex]{\tiny Claude Science, on the same model}};

\node[anchor=north west, font=\sffamily\tiny, text=black!55] at (0.15,-65.6)
  {\faCode\ \ github.com/K-Dense-AI/k-dense-byok \ · \ version 0.9.14 \ · \ macOS, Linux and
   Windows 10/11 \ · \ no programming required};
\end{tikzpicture}

\end{center}

\begin{abstract}
K-Dense BYOK (bring your own keys) is a free, open-source AI research assistant for scientists in any
field that runs on the researcher's own computer.
The researcher supplies access to a model of their choice, hosted or running locally, and the
application supplies everything else: a place for the work to run, a layer of scientific scaffolding,
and a complete record.
Each project is an ordinary folder, so the data, the code, the results, and the record stay on a
machine the researcher administers and can be read years later without the application.
Three things separate it from a chat assistant or a general-purpose coding agent.
It ships a library of written scientific procedures, guided workflow templates, catalogs of where
research data can be found, and reviewer and writer roles the agent can hand work to.
It keeps a Living Lab Notebook whose entries link into an argument and are added to but never erased.
And it records what happened by watching what the agent does rather than by taking the agent's word
for it, in a log the agent has no tool that can write to.
That choice targets the most common failure, model overclaiming, in our earlier benchmark of nine frontier models,
 by making claims checkable rather than preventing them.
On twenty interdisciplinary research prompts, scored under a rubric fixed in advance, K-Dense BYOK led
two managed platforms on both scientific quality and research execution.
Its deliverables were the only ones that recorded the software they ran in, and the only ones that
usually arrived with a command that regenerates the results.
One of the managed platforms ran the same frontier model and supplied neither.
Those environment records were files the agent wrote, not part of the observed log, which does not yet
capture the software environment itself; we state that gap and the others plainly.
The code is available under the MIT license at \url{https://github.com/K-Dense-AI/k-dense-byok}.
\end{abstract}

\clearpage

\section{Introduction}

AI agents, models that take actions such as running code and reading files rather than only answering
questions \citep{Wang2024agentsurvey}, have moved from demonstrations into working scientific pipelines.
They have generated hypotheses for drug repurposing \citep{Gottweis2026coscientist}, guided
iterative rounds of wet-lab experiments by proposing them and interpreting the results
\citep{Ghareeb2026robin}, and built complete analysis products such as transcriptomic aging clocks
\citep{Agarwal2025dendroforge}, and systems that pursue an entire line of inquiry with little human
involvement have begun to appear \citep{Lu2024aiscientist,Mitchener2025kosmos}.
Systems of this kind are often called AI co-scientists, a label that began as the preprint title of
the best known of them \citep{Gottweis2026coscientist}, and a perspective on the area anticipates that
they will progress from tools to collaborators in biomedical discovery \citep{Gao2024biomedagents}.
Almost all of them reach a working scientist as a managed platform, a service in which the vendor
supplies both the model and the software around it and keeps both under its control. That
packaging makes several decisions on the researcher's behalf.
The vendor picks the model, often without saying which one it is, and every request to that model
goes to the vendor, so it cannot be swapped for one that stays inside the institution.
In most such platforms the work itself also runs on the vendor's computers, where the researcher
cannot look, and the data must leave the institution as a condition of use.
A few now run the analysis on the researcher's own machine, Claude Science among them, but whatever
the model reads, including the contents of files, still travels to the vendor with each request
\citep{Anthropic2026claudesciencedocs}.
And the record of what happened is whatever the interface chooses to show.
For someone answering a quick question these are acceptable trade-offs.
For work that has to be defended, reused, or reproduced, each one takes away something a scientist
would normally control.

A second family of tools gives much of that control back: general-purpose coding agents that run on
the user's own computer, use the user's own model access, and work on ordinary files
\citep{Yang2024sweagent,Wang2024openhands}.
What they lack is scientific structure.
They have no notion of a study design, no library of procedures from the scientific disciplines, no
knowledge of where biomedical or earth-science data can be found, no reviewer roles, and no research
record beyond a chat transcript and a list of the commands that were run.
A researcher who adopts one gets control but loses scaffolding; a researcher who adopts a managed
scientific platform gets scaffolding but loses control.

Neither option addresses the failure we find most often in systems of this kind.
In an earlier benchmark, K-Bench~01, we ran nine frontier models, the most capable models available,
through one fixed setup on real scientific requests drawn from users, and had three AI judges score
the results without being told which model produced each answer \citep{kbench01}.
No model reached the level at which a scientist in the field would accept the work with minor edits under all three judges; individual judges placed zero, one, or two models at or above that threshold.
The most common single failure was overclaiming, on 31.4\% of assessments, and in every one of the
nine models both the scientific accuracy and the quality of the delivered files scored below the
quality of the writing.
Writing well and being right are therefore separable, and a system that writes better than it
delivers needs whatever grounds, checks, and documents its claims.
None of that comes from the model.
Human computational science has the same weakness, and it is well measured
\citep{Stodden2018policy,Pimentel2019notebooks,Trisovic2022codequality}.
Of 10,388 published biomedical analysis notebooks whose declared dependencies could be installed,
\citet{Samuel2024jupyter} found that 1,203 ran to completion without error and 879 reproduced the
original result, with missing or unimportable software libraries the largest single cause of failure.
Missing software and incomplete records can prevent reproduction before the scientific reasoning can even be checked.
Practical remedies therefore emphasize recording the steps, software, and inputs needed to regenerate a result \citep{Sandve2013tenrules,vanKampen2024encore}.
Work produced by an agent inherits this problem by default, and faster.

K-Dense BYOK is our attempt to refuse the trade and to treat the record as part of the product
\citep{kdensebyok}.
It is free and open source and runs on the researcher's own computer.
It adds three things to what a general coding agent provides.
The first is scientific scaffolding: written procedures from the disciplines, guided workflow
templates, catalogs of where research data can be found, and reviewer and writer roles the agent can
hand work to.
The second is a Living Lab Notebook, in which the agent records what it thought it was doing as
linked entries that are added to but never erased.
The third is a record of what actually happened, built by watching every action the agent takes and
fingerprinting every file that action touched.
The last of these is the design decision the overclaiming result argues for: a record the agent could
author would not be evidence about the agent, and a model's own account of its reasoning is known to be
unreliable \citep{Turpin2023unfaithful,Chen2025reasoning}.
Software of this kind moves outcomes on its own.
In a study of three models and three software configurations on 100 software-repair tasks, changing only the software shifted success rates by 8.5 to 13 percentage points, against 2.5 to 5 points for changing only the model, and reversed model rankings in six of nine pairwise comparisons \citep{Zhang2026harness}.
Our own earlier system reached 34.4\% on an open-ended bioinformatics benchmark
\citep{Mitchener2025bixbench} while running a model,
Gemini 2.5 Pro, that scores 18.3\% when queried directly \citep{Li2025kdenseanalyst}.
Such claims have to be shown, so we also compared K-Dense BYOK with two managed scientific platforms
on twenty research prompts, and we report both the result and its limits, the largest being that we
ran the evaluation ourselves.

This paper describes what we built, how to start using it, and how it performed.
Section~\ref{sec:design} states the design commitments and the structure that carries them out.
Section~\ref{sec:capabilities} describes what the system can do, with the record of what happened and
the Living Lab Notebook covered at length because they are the parts that address the overclaiming
failure directly, and closes with what is needed to install it and where it is and is not a good fit
today.
Section~\ref{sec:results} reports the twenty-prompt comparison, Section~\ref{sec:discussion}
interprets it, and Section~\ref{sec:limitations} states what the system does not do, including the
parts of its own reproducibility story that are not yet built.
The evaluation protocol is given in Section~\ref{sec:methods}.

\section{Design commitments and architecture}
\label{sec:design}

\begin{figure}[!t]
\centering
\begin{tikzpicture}[
  x=1mm, y=1mm, font=\scriptsize, execute at begin node={\hyphenpenalty=10000\relax},
  every node/.style={text=ink},
  box/.style={draw=black!60, fill=white, rounded corners=1.5pt, align=center,
              inner sep=2.5pt, line width=0.5pt},
  chip/.style={draw=black!45, fill=white, rounded corners=1pt, align=center,
               inner sep=1.5pt, minimum width=25mm, minimum height=8mm},
  flow/.style={-{Stealth[length=2mm,width=1.6mm]}, draw=ink, line width=0.6pt},
  both/.style={{Stealth[length=2mm,width=1.6mm]}-{Stealth[length=2mm,width=1.6mm]},
               draw=ink, line width=0.6pt},
  lbl/.style={fill=paper, inner sep=1.2pt, font=\scriptsize, align=center},
]
\draw[dashed, black!50, fill=paper, rounded corners=3pt, line width=0.5pt]
  (0,0) rectangle (160,-118);
\node[anchor=south west, font=\scriptsize, text=black!60, inner sep=1pt] at (1.5,0.8)
  {\faFolderOpen\ \ Your computer · one project is one ordinary folder};

\node[draw=black!30, fill=black!5, rounded corners=2pt, inner sep=0pt,
      minimum width=112mm, minimum height=14mm, anchor=north] (shelf) at (80,-3) {};
\node[anchor=north west, font=\scriptsize\bfseries, text=black!55, inner sep=1.6pt]
  at (shelf.north west) {Scientific scaffolding supplied by K-Dense BYOK};
\node[chip] (c1) at (39.5,-12)  {{\footnotesize\bfseries 163}\\scientific skills};
\node[chip] (c2) at (66.5,-12)  {{\footnotesize\bfseries 326}\\guided workflows};
\node[chip] (c3) at (93.5,-12)  {{\footnotesize\bfseries 229}\\data resources};
\node[chip] (c4) at (120.5,-12) {{\footnotesize\bfseries 21}\\specialist sub-agents};

\node[box, minimum width=30mm, minimum height=24mm] (you) at (19,-35)
  {{\large\faUser}\\[1pt]{\figttl You}\\[2pt]a question and your data\\
   your own model access,\\hosted or local,\\chosen per task};
\node[box, draw=ink, line width=0.9pt, minimum width=52mm, minimum height=24mm] (kady) at (80,-35)
  {{\large\faRobot}\\[1pt]{\figttl Kady, the agent}\\[2pt]
   reads your files · writes and runs code\\searches and reads sources\\
   asks when a requirement is unclear\\delegates to specialists};
\node[box, minimum width=30mm, minimum height=24mm] (res) at (141,-35)
  {{\large\faChartBar}\\[1pt]{\figttl Results}\\[2pt]figures · tables\\report · code\\
   ordinary files you can\\open without the app};

\draw[both] (you.east) -- node[lbl, above=1pt]{question, data}
                          node[lbl, below=1pt]{clarifying\\questions} (kady.west);
\draw[flow] (kady.east) -- node[lbl, above=1pt]{delivers} (res.west);
\draw[flow] (shelf.south) -- node[lbl, right=1pt]{draws on} (kady.north);

\node[draw=nbk, fill=nbk!7, rounded corners=2pt, line width=0.6pt, inner sep=0pt,
      minimum width=70mm, minimum height=36mm, anchor=north west] (nb) at (4,-54) {};
\node[anchor=north west, inner sep=2.5pt, align=left] at (nb.north west)
  {{\figttl\textcolor{nbk}{\faBook}\ \ Living Lab Notebook}\\[-0.5pt]
   {\itshape\color{black!55}what Kady reasoned}};
\foreach \px/\plt/\pnm in {16/H/hypothesis, 32/M/method, 48/O/observation, 64/D/decision} {
  \node[circle, fill=nbk, text=white, font=\tiny\bfseries, inner sep=0pt,
        minimum size=3.6mm] (d\plt) at (\px,-73) {\plt};
  \node[below=0.8pt of d\plt, inner sep=0.4pt] {\pnm};
}
\draw[flow, draw=nbk, line width=0.5pt] (dH) -- (dM);
\draw[flow, draw=nbk, line width=0.5pt] (dM) -- (dO);
\draw[flow, draw=nbk, line width=0.5pt] (dO) -- (dD);
\draw[-{Stealth[length=1.5mm,width=1.2mm]}, draw=nbk, line width=0.5pt]
  (dO.north) to[bend right=30]
  node[above=-0.5pt, font=\tiny, inner sep=1pt, fill=nbk!7, text=nbk]{supports / refutes} (dH.north);
\node[anchor=south, text=black!55, inner sep=0pt] at (39,-87.5)
  {append-only: corrections supersede, never overwrite};

\node[draw=prov, fill=prov!7, rounded corners=2pt, line width=0.6pt, inner sep=0pt,
      minimum width=70mm, minimum height=36mm, anchor=north west] (pv) at (86,-54) {};
\node[anchor=north west, inner sep=2.5pt, align=left] at (pv.north west)
  {{\figttl\textcolor{prov}{\faFingerprint}\ \ Provenance log}\\[-0.5pt]
   {\itshape\color{black!55}what actually happened, one row per tool call}};
\node[anchor=north west, inner sep=0pt] at (89,-63.5) {%
  \begin{tabular}{@{}l@{\hspace{4pt}}l@{\hspace{4pt}}l@{\hspace{5pt}}l@{}}
  \ttfamily read  & \ttfamily counts.csv       & \ttfamily e9c2\ldots & {\tiny\color{prov}observed}\\
  \ttfamily run   & \ttfamily de\_analysis.py  &                      & {\tiny\color{prov}observed}\\
  \ttfamily write & \ttfamily volcano.png      & \ttfamily a1f3\ldots & {\tiny\color{prov}observed}\\
  \end{tabular}};
\node[anchor=north west, inner sep=0pt, align=left, text=black!55] at (89,-75.5)
  {every file hashed (SHA-256); every link labeled\\observed, inferred, or declared};
\node[anchor=south west, inner sep=0pt, align=left] at (89,-87.3)
  {\textcolor{prov}{\faLock}\ \ Kady has no tool that can write here};

\draw[flow] (66,-47) -- node[lbl, left=1pt, align=right]{writes entries\\as it works} (66,-54);
\draw[flow, dashed] (94,-47) -- node[lbl, right=1pt, align=left]{observed by\\the app} (94,-54);

\node[draw=black!40, fill=white, rounded corners=1.5pt, inner sep=2.5pt, align=center,
      text width=118mm, anchor=north] (chk) at (80,-95)
  {a notebook entry cites an artifact by its path and provenance tracks the same artifact by
   its hash, so every cited artifact reads as \textbf{current}, \textbf{stale}, or
   \textbf{unverified}};
\draw[flow] (39,-90) -- (39,-92.5) -- (62,-92.5) -- (62,-95);
\draw[flow] (121,-90) -- (121,-92.5) -- (98,-92.5) -- (98,-95);

\node[rounded corners=3.5pt, fill=ink, text=white, font=\footnotesize\bfseries,
      inner xsep=8pt, inner ysep=3.5pt, anchor=north] (pill) at (80,-107)
  {claims that can be checked \ · \ work that can be re-run};
\draw[flow] (chk.south) -- (pill.north);
\end{tikzpicture}
\caption{How K-Dense BYOK works, from the researcher's side.
The researcher brings a question, data, and access to a model; Kady, the agent, does the work in a
project folder on the researcher's own machine, drawing on the scaffolding shown at the top, and
returns ordinary files.
Two records are written as it works: Kady writes the Living Lab Notebook, and the application writes
the provenance log by observing Kady's actions.
Because both refer to the same files, each file a notebook entry cites can be reported as current,
stale, or unverified.}
\label{fig:arch}
\end{figure}
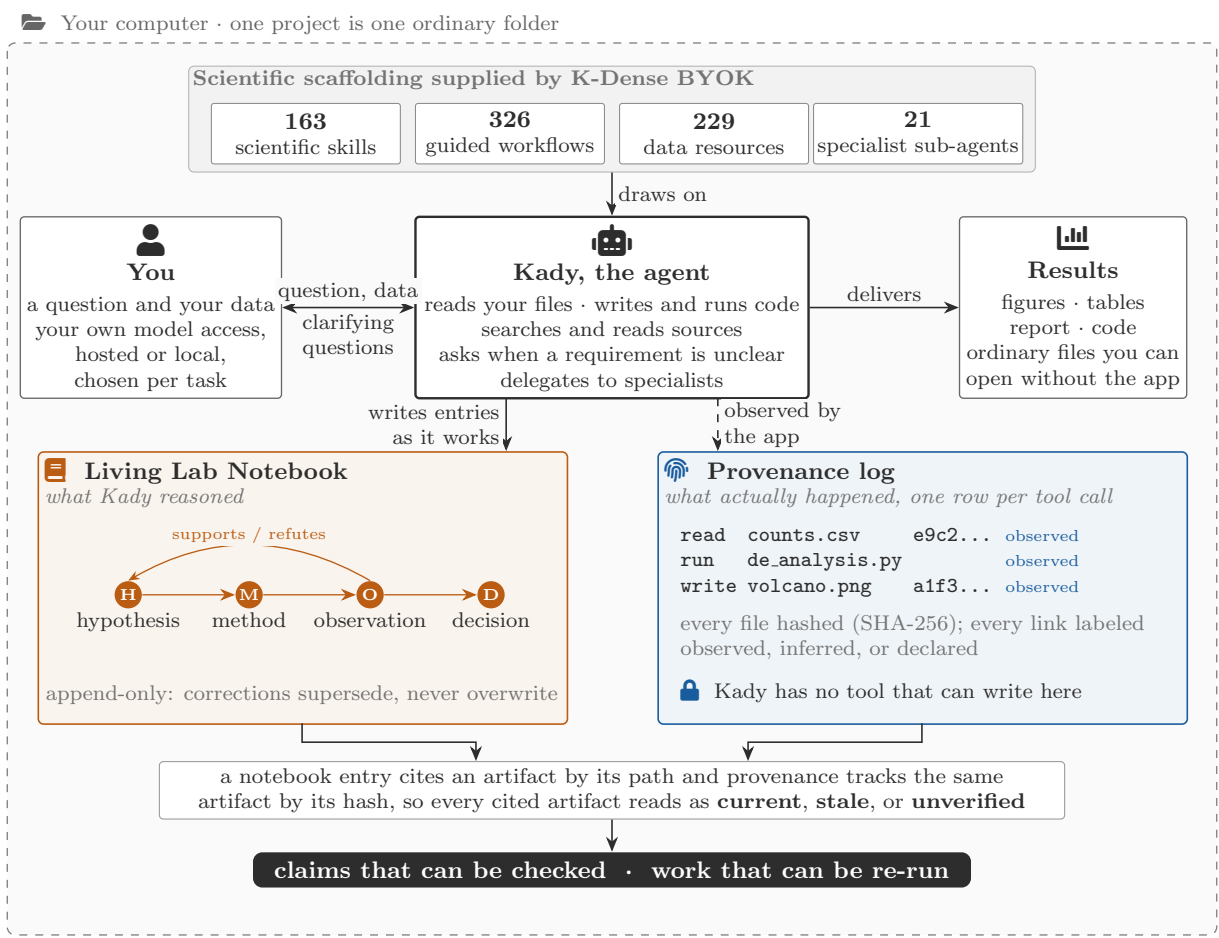

\subsection{Four commitments}

K-Dense BYOK is free, open source under the MIT license, and still in beta; the version described
here is 0.9.14 \citep{kdensebyok}.
Four commitments shaped it.
Figure~\ref{fig:arch} shows how they fit together from the researcher's side.

The first is that the user brings the necessary API keys.
The application supplies the place where work runs, the scientific scaffolding, and the record; the
user supplies access to a model.
Hosted models can be reached through a single OpenRouter account, which gives pay-as-you-go access
to models from many companies, through NVIDIA's hosted models using NVIDIA credits, or through an
existing subscription (ChatGPT Plus/Pro, Claude Pro/Max, GitHub Copilot, or xAI) connected by
signing in to that account from within the application.
Models running on the researcher's own machine, served by free programs such as Ollama or LM Studio,
appear in the same menu as the hosted ones.
The model is chosen for each chat rather than once per installation, so the model and its reasoning
depth can be matched to the task, and a task whose data must not leave the machine can be run
entirely on a local model.
Through OpenRouter, a chat can also put a question to a panel of several models at once and have a
judge model combine their answers into one.
That mode suits questions of interpretation and judgment rather than analysis, because during such a
turn the agent cannot read files or run code, the answer is not exactly repeatable, and the sources
the panel consulted are not shown.

The second is that the workspace is local and ordinary.
Each project is a folder on the researcher's computer.
Inside it is a working folder, which the software calls the "sandbox", where the agent does its
work, and
all artifacts of the work live there as files: conversation history, notebook entries, cost records,
provenance entries, the state of any remote computation, and the results themselves.
We host nothing.
A project can be opened with other software, moved, archived, backed up, or read years later without
the application, which ties the researcher to the vendor far less than a database only the vendor
can open.

The third is that the record comes from observation.
The agent can write files, run commands, and add notebook entries, but it has no tool that writes to
the provenance log.
The application produces that log itself, by watching the same stream of events that drives the
interface.
Provenance is what the model is checked against, so a record the model could write would defeat its
own purpose.

The fourth is that limits are reported rather than hidden.
Where the provenance recorder cannot establish something, it labels the uncertainty; where a limit
is exceeded, the affected step says so.

\subsection{Architecture}

A single start command launches two programs on the researcher's machine.
The first is the interface, which opens in the researcher's web browser but is served from
their own computer rather than from a website.
It provides the chat, project and file management, scientific previews, editors, model-provider
settings, and a view of costs.
The second is a background program that does the work.
It runs the agent, manages its tools and its access to models, keeps each project's working folder,
records what happens, runs specialists, enforces budgets, and oversees long-running computation.
The single agent it runs, Kady, has built-in
tools for reading and writing files and running commands, a tool for handing work to specialists,
and any outside tools connected through the Model Context Protocol \citep{mcp2024}, an open standard by
which outside programs such as reference managers, databases, and lab software can offer their
functions to an AI assistant.
Requests to the model go directly from the agent to the provider the researcher chose, under that
provider's own privacy terms; nothing of ours sits in between that could read or keep what is sent.

Two design choices matter more than the rest.
The first is that the background program, not the browser window, owns a piece of work while it is
running.
Refreshing or closing the browser tab therefore only stops the researcher from watching; the work
continues, and reopening the tab shows what was missed.
This protection ends at the background program itself: restarting it ends any work still in
progress, while completed history and cost records remain on disk.
The second is that each chat tab is an independent conversation, with its own message history, choice
of model, attached files, and cost record, while every tab in a project shares one working folder.
A researcher can therefore run up to ten lines of work at once on the same files, and a file written
by one tab is immediately visible to the others.
Several projects can run at the same time as well, and the project list shows which are working,
finished, waiting for an answer, or stopped by an error.
Heavy computation can be sent to rented cloud machines, with or without graphics processors (GPUs),
through the Modal service, on a path built to survive interruptions.
The application sets aside the job's estimated cost from the project budget before anything starts,
saves a record of the job inside the project, and copies the requested result files back only once
all of them have arrived, so an interrupted transfer never leaves a partial result behind.
Because the record identifies the remote machine, a job still running when the background program
restarts is found again and followed to completion, and every job can be watched, cancelled, or
retried from the workspace.

Python analyses run in an environment that the application sets up and manages automatically inside
each working folder, so the researcher never installs packages by hand, and the catalog of scientific
skills is copied into the project the first time it starts.
Sign-in credentials for model providers are kept in one protected file outside the project folder,
so a specialist running as a separate program can use the same sign-in without the credentials ever
being copied into projects, conversation records, or the commands that start those programs.

\section{Capabilities}
\label{sec:capabilities}

A researcher works with Kady much as they would with a computationally fluent colleague.
Suppose a scientist drags a table of sequencing counts into a new project and types a sentence:
compare the treated and control samples and show me which genes change.
Kady first asks what it needs to know, through a short question form in the chat with suggested
answers, rather than guessing which samples belong to which group or what form the output should
take.
Then it works, and the researcher watches: it reads the file, writes an analysis script, runs it,
notices that a batch effect is confounding the comparison, says so, and adjusts.
It can search the web and read web pages, PDFs, and code repositories along the way, and it can hand
parts of the job to the specialists described below, such as asking the statistics reviewer sub-agent to check
the test it chose.
A researcher can also paste an image, such as a gel photograph or a published figure, into a message
for a model that accepts images.
Key results, including tables, statistical tests, plots, and quality-control checks, appear as
structured cards in the chat that link to the underlying files; these cards organize what the agent
reported and are not an independent check of it.
Follow-up messages can be queued while a task runs, and several chats can proceed at once on the
same project files (Section~\ref{sec:design}).
What is left afterwards enables reproducibility: a folder holding the script, the figures, the tables, a report,
a notebook of what the agent concluded and why, and a record of every action it took.
No programming is required at any stage, but everything the agent writes is an ordinary file the
researcher can open, edit, and run again.

\subsection{Scientific scaffolding}

The scaffolding layer is what sets the system apart from a general coding agent, and it is a
collection of content rather than a piece of machinery (Table~\ref{tab:inventory}).
Skills are written procedures, stored as plain text (Markdown) files in the project in an open format
shared with other agent software \citep{AgentSkills2025spec}, that the agent finds and follows using its
ordinary tools.
They come from Scientific Agent Skills, an openly licensed library that the application installs into
each project and keeps up to date \citep{Kassis2026skills}. Each one records the procedural choices a
defensible analysis rests on: which test the field accepts, which naming scheme is authoritative, and
which caveats a result must carry.
They cover genomics, proteomics, bioinformatics, drug discovery, chemistry, materials science, and
clinical research.
The agent turns to a skill when it is relevant, and a researcher can browse the skills, switch one
off without deleting it, edit one, write a new one in plain language, or install skills that others
have published, from a code-sharing site or a folder on disk, fixed at a chosen version.
The catalog itself is refreshed daily, and a skill the researcher has edited is kept and flagged for
review rather than overwritten.
Guided workflow templates turn recurring analyses into fill-in-the-blank starting points that open
into a chat.
Data-resource entries tell the agent where biomedical, chemical, scholarly, earth-science, climate,
space, and market data can be found.
Specialists are helpers the agent can hand work to.
Each has its own instructions, an optional choice of model, a set reasoning depth, and limits on
which tools it may use.
The roster includes reviewers for statistics, code, methodology, citations, and ethics, auditors for
machine learning and reproducibility, and writers for protocols, abstracts, and manuscripts.
Kady decides on its own when handing off work is worthwhile, or a researcher can name a specialist
directly, and specialists can run one at a time, side by side, or in sequence.
A researcher can also change a specialist's instructions or create a new one, such as a checker for
a particular assay, by describing its job in plain language.

\begin{table}[t]
\centering
\caption{Scientific scaffolding shipped with K-Dense BYOK 0.9.14. The workflow, resource, and
specialist counts were verified against the source repository. The skill catalog is copied into each
project from the Scientific Agent Skills library the first time the project starts
\citep{Kassis2026skills}; that library is versioned separately and keeps growing, so the count given
here is the one current at the time of writing, and the application's own README still reports an
earlier figure of 149.}
\label{tab:inventory}
\small
\begin{tabular}{@{}llr@{}}
\toprule
Layer & Organization & Count \\
\midrule
Scientific skills & procedures followed with ordinary tools & 163 \\
Guided workflow templates & across 22 categories & 326 \\
Data-resource entries & across 18 categories & 229 \\
Specialists & helper agents Kady can hand work to & 21 \\
Previewable scientific formats & structures, spectra, imaging, arrays & 60+ \\
\bottomrule
\end{tabular}
\end{table}

Around the agent sits a workspace built for scientific files rather than program code.
More than sixty scientific file formats open directly in the workspace with a single click,
including interactive three-dimensional protein and molecular structures that can be rotated,
two-dimensional chemical structures, mass spectra, NMR and infrared spectra, chromatograms, sequence
alignments, phylogenetic trees, single-cell and array data, and medical, neuroimaging, and microscopy
images (DICOM, NIfTI, and TIFF), alongside ordinary spreadsheets (CSV), PDFs, text, images, code,
and Jupyter notebooks.
Previews are generated on the researcher's own machine, large files are shown a slice at a time
rather than loaded whole, and patient-identifying fields in medical images are never displayed.
Manuscripts can be written in LaTeX inside the application, with the source and the finished PDF
side by side, the PDF rebuilt automatically as the author works, and AI-suggested edits shown as
marked-up changes that the author can accept or undo.

\subsection{Provenance from observation}

Provenance answers one question about any file the agent produces: where did this come from, and
could it be obtained again?
The question is old, and the workflow community has answered it with standards for recording how a result
was produced and packaging that record so others can read it
\citep{Moreau2013provdm,Khan2019cwlprov,Leo2024rocrate}.
The provenance literature distinguishes a workflow's planned steps from the record of what actually ran \citep{Freire2008provenance}.
A predeclared workflow is not required by all these formats: RO-Crate can also describe individual computations without one \citep{Leo2024rocrate}.
An agent can choose its next steps as it goes, so its record has to follow those decisions.
Tools that record a program's run by watching it, tracing the files it opens and the libraries it loads
rather than asking the author to declare them, are older than agents and show that observation can also
capture the software environment \citep{Murta2015noworkflow,Chirigati2016reprozip}.
Recent work extends provenance models to cover agent decisions and prompts, but captures them from inside
the agent's own program through instrumentation added to its tools and model calls \citep{Souza2025provagent},
and a survey of the area finds no shared scheme for what an agent run should record \citep{Wang2026agenttraces}.
Our departure is to take the record out of the agent's hands entirely, and a recent proposal for
auditable agents points the same way, having a separate layer intercept each action and write a record
of it that cannot be altered unnoticed \citep{Nian2026auditable}.
Every action the agent takes (a tool call, in the agent's terms) becomes one line in a plain-text
log kept inside the project folder, and lines are only ever added, never changed.
Each line records which tool was used and with what inputs, when it ran, which run it belonged to,
which model was in effect, and which files in the working folder the action read and wrote, each
with a
fingerprint of its contents.
The fingerprint is a short code computed from the file's contents (a SHA-256 hash); it changes if
even one character of the file changes, so a matching fingerprint means the contents are identical.
Opening any file in the workspace shows its current fingerprint, the step that produced it, the
inputs that step read, the run and model responsible, and every notebook entry that cites it.

The design is tested by the cases where it cannot be sure which step touched which file, and in
those cases it labels the doubt instead of guessing.
Every link between a step and a file carries a label.
When a tool names the file it works on, as the reading, writing, and editing tools do, the link is
labeled \emph{observed}.
Commands the agent runs are opaque, and running an analysis script is how most real scientific
outputs get made, so only a comparison of the working folder before and after the command can show
what it did.
Links found that way are labeled \emph{observed} too, unless a neighboring step could have absorbed
the change, in which case they are downgraded to \emph{inferred}.
A link the model merely asserted is labeled \emph{declared} and is never upgraded.
Work handed to a specialist is reconstructed afterwards from the specialist's own record, because the
application does not watch it as it happens, so its files are fingerprinted when they are collected,
which may be later than when they were written.
A later match therefore proves only that nothing changed since the record was taken, while a mismatch
remains decisive.

The fingerprints exist mainly to make one hazard detectable.
A notebook entry that cites a figure is a claim about the version of the figure that existed when
the entry was written.
If the figure is regenerated after a bug is fixed, the citation may point at something else
while the surrounding text still reads as though it describes the original image.
The workspace therefore reports each file as \emph{current}, \emph{stale}, or \emph{unverified},
flags each citation that was written before the file's latest version.
Matching file size and modification time are not treated as proof that a file is unchanged.
Where the recorder reaches one of its limits, it marks the affected step instead of quietly dropping
the information (Table~\ref{tab:bounds}).

\begin{table}[t]
\centering
\caption{Limits in the provenance recorder and how they are reported. Each limit marks the affected
step instead of truncating its record unannounced; the values were read from the source code.}
\label{tab:bounds}
\small
\begin{tabular}{@{}lll@{}}
\toprule
Limit & Value & What happens when exceeded \\
\midrule
Files scanned per comparison & 20{,}000 & step marked as too large to compare fully \\
Largest file fingerprinted & 512\,MB & size and time stamp kept; marked not fingerprinted \\
File links recorded per step & 200 & extra links dropped; the number dropped is reported \\
Tool arguments stored per step & 4\,KB & stored as a shortened preview \\
Comparison fails or has no baseline & -- & step marked as failed or as lacking a baseline \\
\bottomrule
\end{tabular}
\end{table}

\subsection{The Living Lab Notebook}

Electronic laboratory notebooks have long been proposed to make the reasoning behind an experiment
recoverable, yet uptake in academic laboratories has lagged, and adoption depends in part on fitting laboratory workflows without adding undue work
\citep{Nussbeck2014notebook,Dirnagl2016eln,Kanza2017eln,Higgins2022eln}.
An agent removes that obstacle, because it can be made to write the entry as a condition of taking the
step.
The provenance log records what the system did; the notebook records what it thought it was doing,
and the two can be linked.
The agent writes entries as the work proceeds, through a tool that does not interrupt it.
Each chat keeps its own notebook, and a project-wide view merges the entries of every chat for
reading.
Each entry has a kind (hypothesis, method, observation, decision, or note) and can carry a written
body, code, a confidence level, tags, and the locations of files in the working folder.
Entries link into an argument rather than a diary.
An observation that tests a hypothesis points at it and takes a stance of supporting, refuting, or
neutral, and the hypothesis then shows a live status of open, supported, or refuted, decided by the
most recent link that is not neutral.
Nothing in the history is ever erased.
Correcting an entry means adding a new one that replaces it; the original stays, struck through,
with links in both directions, so the record keeps the fact that a conclusion changed.
The researcher's own annotations (pins, comments, and notes) are stored in a separate companion
file, so the agent's record is never altered by the human layer.
The notebook can be exported as a plain-text document, as structured data that other software can
read (JSON), or as a single archive that includes every file it refers to, with links rewritten to
work inside the archive, and it can be printed to PDF.
A draft Methods section can be generated from the method, decision, and observation entries.

Two properties make this more than a convenience.
The record is written during the work, while the reasoning is still available, rather than pieced
together afterwards.
And because entries cite files by their location while the provenance recorder tracks those files by
their fingerprints, a citation that has gone stale can be caught rather than trusted.
Together these target the overclaiming failure that dominated K-Bench~01 \citep{kbench01}.
A claim recorded next to the file it rests on, with that file's identity tracked independently, is a
claim that can be checked.

\subsection{Cost tracking and control}

Because the researcher's own account pays for the work, cost is recorded alongside the work itself,
not as a separate billing detail.
Usage, measured in tokens (the units that model providers bill by), and its value at list price are
recorded for each run and each project, and usage that is billed as it goes counts against an optional
firm spending limit for the project.
Remote computation sets aside its estimated cost against the project budget before it runs.
Throughout, the researcher can watch each step, inspect and edit the code, redirect a run in
progress, and stop it.

\subsection{Getting started}

Start with the \href{https://github.com/K-Dense-AI/k-dense-byok#readme}{K-Dense BYOK README}, which contains the most up-to-date installation instructions.
Installing the application takes a few minutes and no programming.
It runs on macOS, Linux, and Windows~10/11, and needs only two free programs that many computers
already have: Node.js version 22.19 or later, which the start command installs on a Mac if it is
missing, and git, the standard program for downloading code.
On Windows, git comes as Git for Windows, which also supplies the command shell the agent uses to run
its work.
The researcher downloads the code from the repository, copies the template settings file, and runs
one start command; everything else, including the scientific packages and the catalog of skills, is
installed the first time it starts.
The interface then opens in the ordinary web browser, served from the researcher's own computer.
The last step is to give it access to at least one model, in any of four ways: an OpenRouter account
key, which reaches models from many companies on a pay-as-you-go basis; a key for NVIDIA's hosted
models; an existing ChatGPT, Claude, GitHub Copilot, or xAI subscription, connected by signing in
from within the application; or a model running on the researcher's own machine.
Two details are worth knowing in advance.
The subscription route works only by signing in, so a plain provider key for the same account is not
accepted in its place, and a few features still require an OpenRouter key.
Video tutorials are available in the \href{https://www.youtube.com/playlist?list=PLNsALmWNTJ5m9fd4CqYK0o06KwSJ6nr86}{K-Dense BYOK YouTube playlist}.

Work whose data cannot leave the institution has its own path.
Free local programs such as Ollama and LM Studio, or any server that speaks the same common
interface, serve models from the researcher's own machine, and those models appear in the same menu
as the hosted ones, so a project can be run without anything being sent to an outside company at all.
Any model those programs offer that supports tool calling, the ability to run commands and read files
rather than only reply, can be used; Gemma~4, Qwen3.8~27B, and Nemotron~3.5 Lightning have each
completed a multi-step analysis end to end in our own use, though we report no evaluation of them here,
and a laboratory with a multi-GPU server can run larger models still.
The trade is capability: in the Berkeley Function Calling Leaderboard, several small models performed well on individual tool calls but poorly on tasks requiring repeated tool use \citep{Patil2025bfcl}.
Following written procedures is another limitation, as Section~\ref{sec:limitations} describes.
Because each chat picks its own model, a sensitive analysis can run locally in one tab while a
literature search runs on a hosted model in another.

\subsection{Where it fits today}

The system is in beta, and some kinds of work suit it better than others.
It fits work where a researcher has data in hand and a question that needs analysis, code, figures,
and a written result: exploratory analysis of an assay or a public dataset, a literature-backed
review with the sources attached, a reanalysis someone else must later be able to repeat, or drafting
a manuscript alongside the files it rests on.
It fits equally where the record matters as much as the answer, which includes anything a reviewer,
a collaborator, or a regulator may later ask about.
It is a poorer fit in two situations.
Very small local models are not yet capable of multi-step tool use.
And several capabilities were set aside during a rebuild of the system's core and are not yet
available, including built-in literature search and a Methods export drawn from the record.
Both are stated in full in Section~\ref{sec:limitations}.
In short, the system is useful now but unfinished.

\section{Evaluation against two managed platforms}
\label{sec:results}

We compared K-Dense BYOK with two managed scientific platforms, Claude Science and Biomni Lab, on
twenty interdisciplinary research prompts covering oncology, immunology, single-cell biology, remote
sensing, climatology, seismology, and other fields.
Each platform produced one bundle of deliverables per prompt, from a single submission of the prompt
with no follow-up messages, and all three completed all twenty.
All sixty bundles were produced and scored in July 2026.
The K-Dense BYOK runs therefore used a build from seven weeks and sixteen releases before the version
described in this paper.
Both comparison platforms are vendor-controlled products that change continuously, so what follows
describes them as they behaved in that window rather than as they behave now.
Biomni Lab runs entirely in the vendor's cloud and exposes no version number, and we did not record
the version of Claude Science we ran.
Bundles were checked twice: first by an automated inventory of what each contained, then by a single
judge model scoring them under a rubric.
For these runs we selected Claude Opus~4.8 on both K-Dense BYOK, at its highest reasoning level, and
Claude Science, so that comparison holds the model roughly fixed and varies the rest of the system.
Biomni Lab's Max profile does not name the model behind it, and Phylo states that it matches models to
tasks, so its model may differ between prompts \citep{Phylo2026modelscompute,Phylo2026modelagnostic}.
We designed, ran, and scored the evaluation ourselves, which is the study's main weakness and is set
out with the others in Section~\ref{sec:limitations}.
The full protocol is in Section~\ref{sec:methods}.

\subsection{What the deliverables contained}

\begin{figure}[t]
\centering
\includegraphics[width=0.9\textwidth,trim=0 21 0 58,clip]{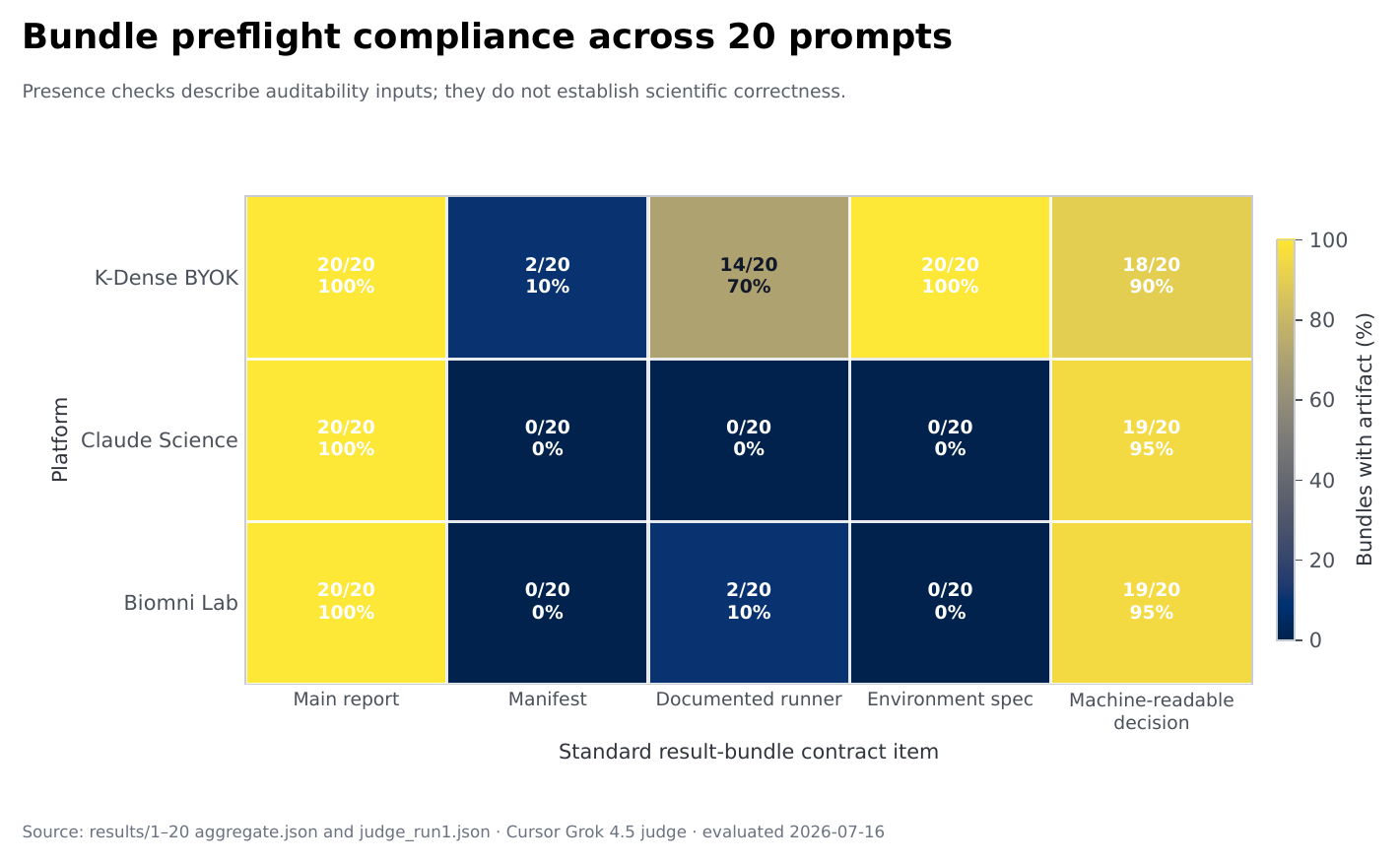}
\caption{What each platform's deliverables contained, over the twenty prompts, for five of the seven
items on the standard list.
Cells show how often each platform's bundles contained each item.
These checks measure only what is needed for an audit and do not establish scientific correctness.}
\label{fig:preflight}
\end{figure}

Before any scoring, an automated check inventoried every bundle against a standard list of seven
items: the main report, a list of files with their fingerprints, a runnable command that regenerates
the results, a record of the software environment, results in a form other software can read,
receipts for the sources used, and logs of what was run.
Most of these items are long-standing recommendations for reproducible computational work
\citep{Sandve2013tenrules}.
These are yes-or-no checks for whether an item is present and involve no judgment.
Figure~\ref{fig:preflight} reports five of the seven; source receipts and run logs were assessed only
by the judge, under the criteria that cover them.
On the items that come from reporting alone, the main report and the machine-readable results, the
three platforms cannot be told apart.
The difference is confined to the items that let someone else reproduce the work.
K-Dense BYOK included a record of the software environment on every prompt and a documented
regeneration command on most.
Biomni Lab included a regeneration command on a few prompts and neither of the other two items.
Claude Science included none of the three on any prompt.
Lists of files with fingerprints were rare everywhere, K-Dense BYOK included.
Their presence does not certify that the work is correct, but their absence makes a bundle impossible
to audit.

The environment records that earned credit here were files the agent itself produced in the bundle,
helped by the Python environments the application manages in the working folder.
More generally, the inventory and the rubric score a bundle as a collaborator would receive it, and
neither was designed around K-Dense BYOK's provenance log or notebook; the inventory did not count the
file fingerprints in that log as a manifest.
The evaluation therefore measures what the scaffolding and the working folder led the agent to
produce, not the record described in Section~\ref{sec:capabilities}, and the two should not be
confused.

\subsection{Judged outcomes}

\begin{figure}[!htbp]
\centering
\includegraphics[width=0.94\textwidth,trim=0 21 0 60,clip]{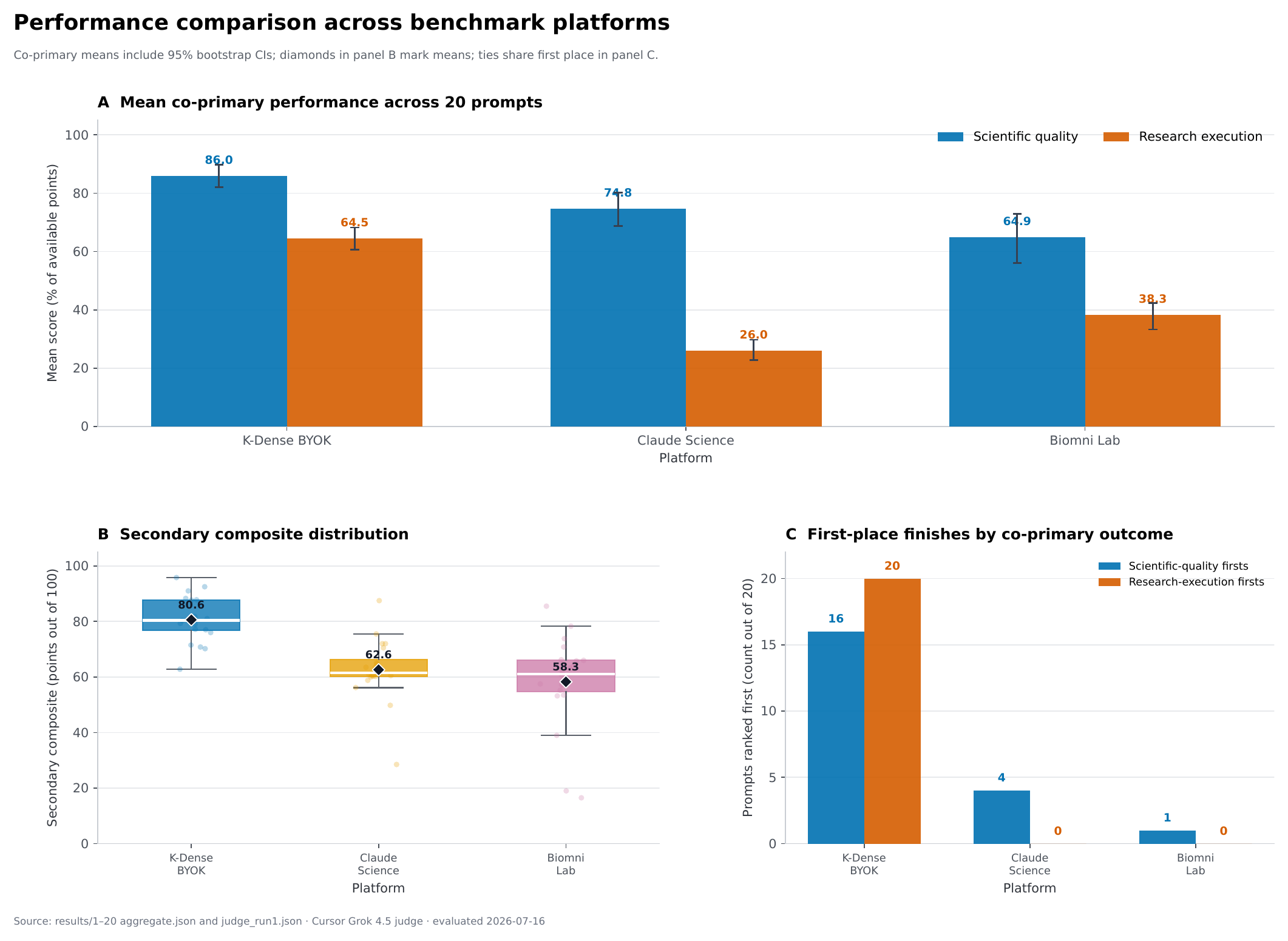}
\caption{Summary of platform performance.
(A) Mean outcomes with 95\% bootstrap confidence intervals over prompts.
(B) Distributions of the secondary 0--100 combined score, with diamonds marking platform means.
(C) Number of first-place finishes on each outcome; tied platforms share first place.}
\label{fig:summary}
\end{figure}

\begin{figure}[!htbp]
\centering
\includegraphics[width=0.85\textwidth,trim=0 21 0 32,clip]{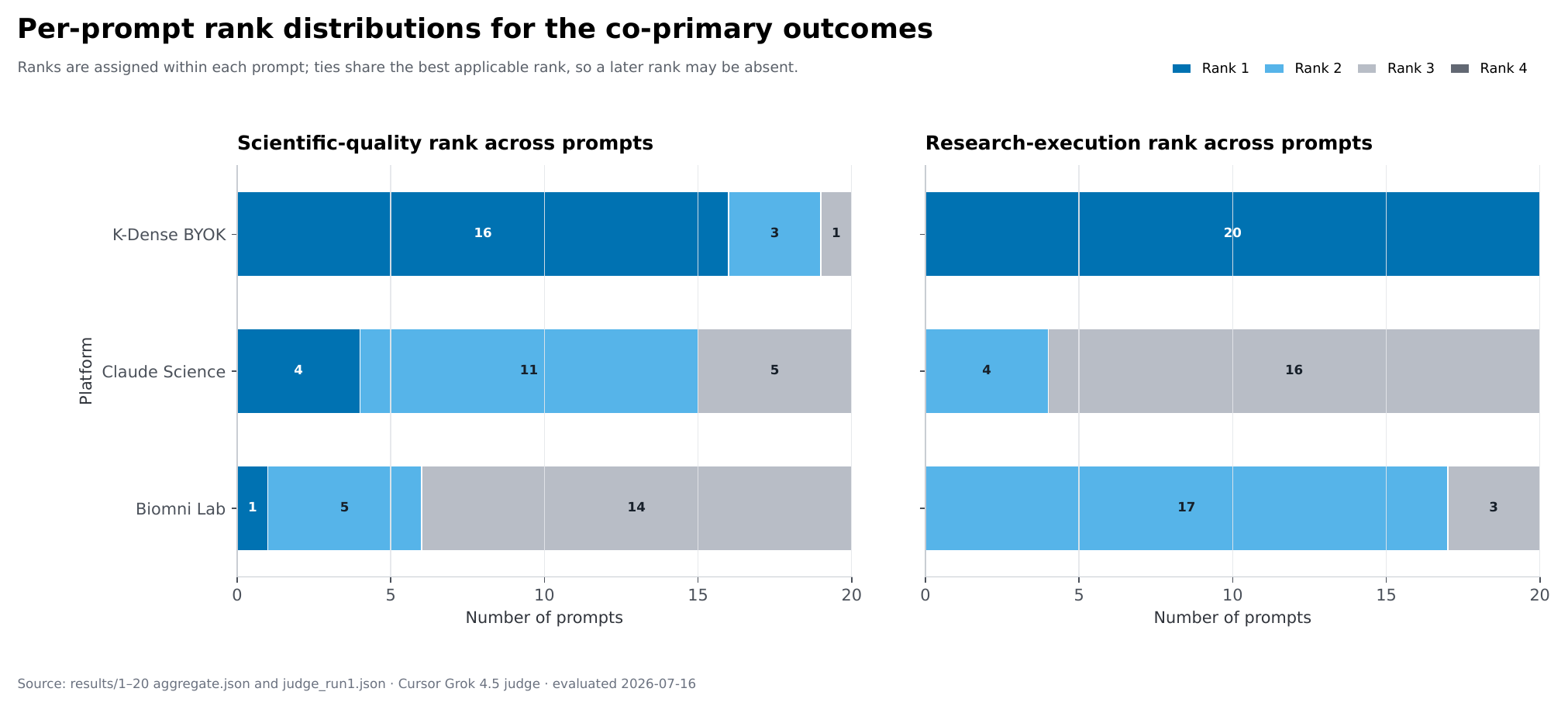}
\caption{How often each platform placed first, second, or third on the two outcomes.
Bars count prompts; ranks are assigned within each prompt, and tied platforms share the better rank,
so a lower rank may go unused.
On research execution K-Dense BYOK ranked first on all twenty prompts and neither comparison
platform ranked first on any.}
\label{fig:ranks}
\end{figure}

The rubric scores each bundle on two outcomes: scientific quality,
worth 75 points across nine criteria, and research execution, worth 25 points across four criteria that
cover whether the work can be regenerated, traced, and audited.
K-Dense BYOK led on both (Table~\ref{tab:means}), but the two outcomes tell different stories.
On research execution the spread is far wider than on scientific quality: Claude Science falls to
last, and Biomni Lab takes second place.
Figure~\ref{fig:summary} shows the same comparison three ways: the means with their intervals, the
spread of the combined score, and how often each platform finished first.
Plotted against each other, the two outcomes separate K-Dense BYOK from both comparison platforms
almost entirely along the execution axis (Figure~\ref{fig:landscape} in
Appendix~\ref{app:figures}).
The execution lead also has no exceptions (Figure~\ref{fig:ranks}).
K-Dense BYOK scored highest on execution on every one of the twenty prompts. On scientific quality each comparison platform took first
place on at least one prompt.
Appendix~\ref{app:figures} gives the per-prompt scores and the criterion profile behind these
summaries.

\begin{table}[t]
\centering
\caption{Judged outcomes over the twenty prompts, as \% of available points (scientific quality out
of 75, research execution out of 25), with 95\% bootstrap confidence intervals obtained by resampling
prompts. The lower rows give the paired difference by prompt, in percentage points, between K-Dense
BYOK and each comparison platform, and the number of prompts on which K-Dense BYOK scored higher than
that platform. Intervals are recomputed from the per-prompt scores in Table~\ref{tab:perprompt}
(Section~\ref{sec:methods}).}
\label{tab:means}
\small
\begin{tabular}{@{}lcc@{}}
\toprule
 & Scientific quality & Research execution \\
\midrule
K-Dense BYOK & 86.0 [82.2, 89.9] & 64.5 [60.6, 68.2] \\
Claude Science & 74.8 [68.7, 80.0] & 26.0 [22.7, 29.7] \\
Biomni Lab & 64.9 [56.1, 72.7] & 38.3 [33.4, 42.4] \\
\midrule
Difference vs.\ Claude Science & $+11.2$ [6.2, 17.2] & $+38.5$ [33.2, 43.5] \\
\quad prompts led by K-Dense BYOK & 16/20 & 20/20 \\
Difference vs.\ Biomni Lab & $+21.1$ [13.2, 30.1] & $+26.2$ [21.9, 30.9] \\
\quad prompts led by K-Dense BYOK & 18/20 & 20/20 \\
\bottomrule
\end{tabular}
\end{table}

Breaking the scores down by criterion locates the gap precisely
(Figure~\ref{fig:criteria} in Appendix~\ref{app:figures}).
Across the nine scientific-quality criteria K-Dense BYOK leads consistently but by modest margins.
Every platform scores better on how it communicates than on the substance underneath, which
echoes the same pattern we observed in K-Bench~01 \citep{kbench01}.
On the execution criteria the picture changes sharply, and the separation is concentrated in two of
the four.
On both, the mean ratings for Claude Science are near zero and those for Biomni Lab are low. K-Dense BYOK's are the only ones near or above the middle of the scale.
This is a binary that a system either builds and records or does not; they do not follow from reasoning
well about the science.

\section{Discussion}
\label{sec:discussion}

What the comparison shows a working scientist is narrower than a leaderboard and more useful.
Three deployed systems, given the same twenty prompts, were separated most cleanly not by the
scientific quality of their work but by how usable their deliverables were afterwards.
Against the platform running the same frontier model, the lead on quality was modest and the lead on
execution more than three times as large; against the third platform both leads were wide, but only
the execution lead held on every prompt.
That execution gap was concentrated in whether the bundle came with a command that regenerates the
results and a record of the software they ran in, and the automated inventory found the same thing
the judge did.
That distinction is invisible to ordinary quality assessment.
A well-argued report can still be difficult to reproduce if it lacks code, data, or a record of the software environment, problems documented in studies of human computational science \citep{Stodden2018policy,Samuel2024jupyter}.
A researcher choosing a tool should therefore ask what arrives at the end of the session, not only
how the answer reads.

The most informative detail is that Claude Science ran the same frontier model as K-Dense BYOK.
Reasoning well about resistance mutations does not by itself cause a file list to exist or a
regeneration command to be written down.
Something in the software around the model has to produce those and then put them in the researcher's
hands, and where that does not happen the shortfall shows up in the record rather than in the prose.
Anthropic documents that Claude Science keeps the code and the environment behind each figure inside
the application \citep{Anthropic2026claudescience,Anthropic2026claudesciencedocs}, so what we measured is a gap in what the
deliverable carried and not evidence that no such record was ever built
(Section~\ref{sec:limitations}).
Other studies find the same: the software wrapped around a model, not only the model, sets what gets
accomplished.
Holding the model fixed and changing only the interface through which it acts on files and commands
raised the solve rate on a software-repair benchmark from 11.0\% to 18.0\%, a relative gain of 64\%
\citep{Yang2024sweagent}, and the open systems that followed treat that layer as a designed component
of the system \citep{Wang2024openhands}.
Independent studies find the same sensitivity on software-repair tasks and on broader office,
data-analysis and tool-use workflows \citep{Zhang2026harness,Yao2026harnessbench,Zheng2026clawswebench}.
The effect is not automatic, and \citet{Wang2026harnessevolution} show that evolving such software
automatically often fails to beat simply giving the model more computation at run time, which is why
we measured it.
Scientific tasks make the question harder still, because they are open-ended and often have no
single correct answer to check against \citep{Abram2026evalframeworks}.
Existing scientific benchmarks accordingly ask whether an agent can reproduce someone else's work,
whether a published analysis graded against its reference program \citep{Chen2024scienceagentbench},
a study from its published code and data \citep{Siegel2024corebench}, or a paper from its text alone
\citep{Starace2025paperbench}.
The property measured here is the complementary one: whether the agent's own work can be reproduced.

The record built from observation does not yet capture the software environment, so it cannot on its
own tell a reader what would be needed to run the work again (Section~\ref{sec:limitations}).
That is a real contribution and a partial one, and closing the environment gap is the most important
piece of remaining work.
Read alongside K-Bench~01, where no model inside one fixed setup reached the acceptance threshold under all three judges and overclaiming was the leading failure \citep{kbench01}, the practical lesson is that grounding a claim in a record is a different problem from making the claim well, and
only the first is a problem a tool can solve for you.
For the same reason, evaluations of scientific agents should treat the software around the model as
part of the experimental condition rather than crediting the model alone.

The broader argument is about where control should sit.
A researcher using K-Dense BYOK chooses the model for each task, and can choose one that runs on
their own hardware so that nothing leaves the machine, keeps the data on a machine they administer,
reads the record with ordinary tools, and can inspect or change every layer because the system is
open source.
Those properties do not make the science correct, but they make it checkable, and checking comes
first.
A managed platform can offer some of them, and Claude Science now runs its analysis on the
researcher's own computer, but the model stays the vendor's, every request to it goes to the vendor,
and the software in between is closed.

\section{Limitations}
\label{sec:limitations}

\subsection{Limitations of the system}

The system's own reproducibility story has a gap at its center.
The provenance log records what ran but not the environment it ran in, because the versions of
software libraries, the version of the Python interpreter, and random seeds are not captured.
A step therefore tells a reader which tool acted on which file contents, not what would be needed to
reproduce the result, and closing that gap comes first among the remaining work.
The log also does not record the version of the application or of the skill library that was in
effect.
Several narrower limits on provenance follow from the decision to record by observation.
The before-and-after comparison of the working folder runs in the background, so when two actions
finish before the first comparison runs, the changes cannot be split between them and the links are
marked inferred rather than guessed.
A command that starts and finishes before the initial survey of the working folder is complete can
have its writes folded into that survey and go unrecorded.
Changes are detected by file size or modification time, so a rewrite that preserves both goes
unnoticed, though whatever is reported is identified exactly because changed files are
fingerprinted.
Work handed to a specialist is collected only one level deep, so a specialist that itself hands off
work produces a session the parent never learns about, a limit the notebook shares.
Steps run as remote computation are not yet recorded in provenance at all.
And because provenance watches the effects on the working folder rather than the commands
themselves, what a command did internally remains unknown.

The scaffolding depends on the model the user supplies.
Skills are procedures a model must choose to follow, and models sometimes skip a relevant skill,
apply only part of it, misread a multi-step procedure, drop or garble the instructions they send to
tools, follow instructions less reliably in long inputs, as experiments with added irrelevant text have shown \citep{Levy2024sametask}, or
drift from a requested output format.
These effects are strongest with models below the frontier and with small local models, which
struggle most to coordinate several tools \citep{Patil2025bfcl}.
This is a real consequence of the bring-your-own-key design: the application cannot guarantee
behavior it does not supply, and any claim about capabilities in this paper should be read as
depending on a capable model.
A provider's own safety rules travel with its model as well.
At the time of writing, one recent model, Claude Fable~5, refuses any request whose instructions
include certain skills about biological design or pathogens, seven of the catalog, before doing any
work and even when the conversation itself contains nothing sensitive.
The run then fails at once, nothing is billed, and the error names the skills responsible so the
researcher can switch them off or choose another model, but the application cannot override the
policy, and the list of affected skills will change as providers adjust their filters
\citep{kdensebyok}.
The skill library itself carries a matching caveat: it is published as a resource, with structural and
security checks but no task-level evaluation showing that installing a given skill improves an agent's
scientific work \citep{Kassis2026skills}.
The present study measures whole platforms and so does not isolate the contribution of the skills
either.

Two boundaries matter for security.
The agent runs commands under the researcher's own user account, with the same permissions the
researcher has, so the operating system will not stop it from reading the researcher's own saved
credentials.
Telling the agent not to look at secrets is not the same as preventing it; a malicious document
could instruct it otherwise in an attack known as indirect prompt injection
\citep{Greshake2023injection,Debenedetti2024agentdojo}. The agent should run with only the access
it needs \citep{BeurerKellner2025designpatterns}, which on a personal computer means handling untrusted
material in an isolated environment such as a container, a virtual machine, or a separate user account.
Installed skills are instructions rather than data, so installing one written by a third party
extends that trust to whoever wrote it \citep{Schmotz2026skillinject,Liu2026maliciousskills}.
Installing a skill requires the researcher's explicit acknowledgement. Skills installed from a
third party are never updated automatically and their contents are not audited.
Beyond these, subscription allowances cannot be read from inside the application, so a successful
login does not mean usage is free or unlimited, and the remote-computation cost shown is an estimate
rather than an invoice.
Specialists cannot yet use tools connected from outside programs; only Kady can.
Several capabilities set aside while the system's core was rebuilt are not yet available: built-in
literature and regulatory-document search, document conversion, automated web browsing, automatic
citation checking, and a Methods export that draws on the record of what happened.
The citation reviewer among the specialists still works; what is missing is the check that runs
without being asked.
The Methods draft that the notebook already produces is a separate feature, written from notebook
entries rather than from the record of what happened.

\subsection{Limitations of the evaluation}

The first limitation is that a platform bundles a model together with all the software around it,
so the two cannot be cleanly separated.
Holding the model roughly fixed against Claude Science strengthens the case for attributing the
difference to that surrounding software, but it does not make the study a controlled experiment.
The comparison against Biomni Lab remains fully confounded; each
platform still comes with its own settings and rules about tool use; and we did not run a common set
of models through each platform, so the study cannot apportion the variation between the model and
everything else.
The attribution rests on the argument from the criteria, namely that the criteria which separate the
platforms measure files the surrounding software produces, together with the judge-independent
inventory of what each bundle contained, not on direct experimental manipulation.
Running a common set of models through each platform is the obvious confirmatory experiment.

Second, the evaluation used a single AI judge (Grok~4.5) with exactly one
run per bundle, each starting with no memory of the others, and the platform's identity could be
inferred from the submitted files, so the judge was not blinded.
The judge also worked inside a code editor (Cursor) on an author's own computer rather than in a fresh,
isolated environment, and
the same machine served every platform.
Judges of this kind agree well with human raters in aggregate \citep{Zheng2023mtbench,Zhuge2025agentasjudge}
but show meaningful run-to-run noise and systematic biases \citep{Zheng2023mtbench,Feuer2025judgmentnoise}.
One such bias is a preference for a judge's own outputs \citep{Wataoka2024selfpreference}, which we
limited by choosing a judge from a maker whose model, to our knowledge, no evaluated platform ran in
this comparison.
The same work traces the bias to a preference for text the judge finds familiar, which changing the
maker does not remove.
The paired design and the agreement with the presence checks reduce but do not remove the
possibility that some results are artifacts of the judge, and we have not quantified variation
between judges or between runs.

Third, our unit of assessment is the submitted bundle, not the platform's own interface, and for one
platform this distinction is material.
Anthropic states that Claude Science records the code and environment behind each figure and gives every
output an auditable history \citep{Anthropic2026claudescience}, and its documentation describes a
provenance record with an environment tab listing the language version and every installed package
\citep{Anthropic2026claudesciencedocs}.
Our preflight check found no environment record in any of its twenty bundles, and the judge scored it
near zero on the two criteria that depend on one.
Both can hold at once if that record is retained inside the application but not carried into the exported
deliverables, and we did not test the interface to find out.
What reached us as the deliverable could not be regenerated or
audited from its own contents.
Whether a record that a collaborator never receives serves the purpose is a question about deliverables
rather than about the platform's internals, but the distinction should not be read as a measurement of
what the vendor retains.

Fourth, twenty prompts is a modest sample, and the weights the rubric assigns to its criteria reflect
our judgment about what matters rather than values fitted to data; we report the unweighted profile
of criteria so readers can apply their own weights.
The prompts were themselves written with the help of a language model, and we did not test whether
prompts written that way favor any platform.

Fifth, we did not test the other family of tools this paper sets itself against.
A general-purpose coding agent running the same model without the scientific scaffolding is the
condition that would separate the contribution of the scaffolding from the rest of the application,
and we did not run it, so the argument made in the introduction against that family rests on what
those tools do not ship rather than on a measurement.

Finally, and most importantly, the evaluation was not independent: the authors develop K-Dense BYOK
and are employed by K-Dense Inc., a direct conflict of interest.
The safeguards described in Section~\ref{sec:methods} do not substitute for independent replication,
and the results should be read as an internal evaluation under the tested setup rather than a
universal measure of platform performance.

\section{Methods}
\label{sec:methods}

\subsection{Task set}

The benchmark consists of twenty interdisciplinary scientific research prompts, written with the help
of GPT-5.6~Sol and drawn from a broader internal K-Dense benchmark.
They follow the observation in K-Bench~01 \citep{kbench01} that real requests arrive underspecified,
and were written to be as loosely specified as real ones are.
The prompts cover oncology, immunology, pharmacogenomics, single-cell biology, digital pathology,
remote sensing, marine ecology, climatology, hydrology, seismology, and energy equity.
Each requires the agent to obtain data, carry out analyses rather than merely propose them, and
deliver outputs that bear on a decision.
The full text of every prompt is reproduced in Appendix~\ref{app:prompts}.

\subsection{Rubric and judging protocol}

Scoring used rubric version~3.0, a frozen instrument whose criteria are audited against evidence,
with two co-primary outcomes.
Scientific quality is worth 0--75 points across nine weighted criteria (A--I).
Research execution is worth 0--25 points across four weighted criteria (R1--R4) covering executable
regeneration, the lineage of sources, environment and determinism, and traceability.
A secondary combined score sums the two to 100.
Each criterion receives a whole-number rating from 0 to 4, which is converted to weighted points in
proportion.
The rubric credits only evidence present in the submitted bundle or verifiable from the sources it
cites; work that is described but not documented earns nothing, and the judge recomputes key
quantities and attempts to regenerate results where the bundle allows it.
The judge was Grok~4.5.
Each bundle (one per prompt and platform) was judged on its own in a single run, with the judge
starting from a blank slate each time, and all sixty bundles were judged in one pass in July 2026.
Before scoring, the judge extracts a checklist of requirements from the prompt and applies the
identical checklist to all platforms on that prompt.
The judge could tell which platform produced each bundle, because that can be inferred from the
files themselves; this is disclosed as a limitation above.
The full rubric is reproduced in Appendix~\ref{app:rubric}.

\subsection{Platforms and required deliverables}

K-Dense BYOK is the open-source bring-your-own-key application described in Sections~\ref{sec:design}
and~\ref{sec:capabilities} \citep{kdensebyok}: the user supplies access to a model and the application
supplies the place where work runs, the scientific scaffolding, and the record.
For these runs it was set to use Claude Opus~4.8 at its highest reasoning level; the build used was
the one current in mid-July 2026, and its version number was not recorded (Section~\ref{sec:results}).
We selected the same model in Claude Science, so the comparison between the two holds the model
roughly fixed and varies everything around it.
Anthropic states that Claude Science is an application rather than a model and uses the Claude models
included in the user's plan \citep{Anthropic2026claudescience,Anthropic2026claudesciencedocs}, so on both platforms the model was
our choice rather than a property of the platform.
Claude Science was run with delegation, automatic review, and memory enabled, and with no
remote-computation account connected.
These are per-session or per-plan choices, not fixed defaults \citep{Anthropic2026claudesciencedocs},
so we report them as our configuration; we did not record the application version.
Biomni Lab, the hosted platform operated by Phylo \citep{Phylo2026announcement} and descended from the
open-source Biomni agent \citep{Huang2026biomni}, was run under its Max model profile, which its
documentation describes as being for the most demanding reasoning and multimodal work
\citep{Phylo2026modelscompute}.
Each platform produced one frozen bundle of results per prompt.
Every prompt was submitted once, as a single message, with no follow-up messages, steering, or
retries on any platform; each platform was set to the highest reasoning setting it offered; and no
bundle was chosen from among several attempts, because there was only ever one.
The bundle was the complete output folder the platform created for the run, taken as the platform
left it.
An automated preflight check compared every bundle against a standard list of required items: a main
report with a documented regeneration command, a list of files with their fingerprints, a runnable
script or command, a record of the software environment, machine-readable results for the
decision-critical quantities, receipts for the sources used, and logs of what was run.

\subsection{Statistical analysis}

The primary quantities we estimate are the platform means of the two co-primary outcomes, as
percentages of available points.
Uncertainty is expressed as 95\% bootstrap confidence intervals obtained by resampling prompts
\citep{Efron1993bootstrap}, which respects the paired structure in which every platform answers every
prompt.
Means and paired differences are computed from the unrounded scores.
The intervals in Table~\ref{tab:means} were recomputed for this paper from the per-prompt scores in
Table~\ref{tab:perprompt}, which are rounded to whole percentages, using 10{,}000 resamples of the
twenty prompts drawn with replacement, the 2.5th and 97.5th percentiles of the resampled statistic, and
a fixed random seed of 2026, so that a reader can regenerate them from that table alone; the whiskers in
the figures come from the original run and differ from them by at most 0.2 points.
Differences between platforms are reported as paired mean differences by prompt
\citep{Miller2024errorbars}, together with the number of first-place finishes on each co-primary
outcome.
Because the two co-primary outcomes answer different questions, we judge which platform is better on
both outcomes together rather than on the combined score alone.

\section{Conclusion}

A scientist who installs K-Dense BYOK gets a research assistant that works on their own files, on
their own computer, and leaves behind a folder anyone can open: the data, the code, the figures, a
notebook of what it concluded and why, and a record of every action it took.
Two things are worth knowing before starting.
The scaffolding is followed by the model the researcher supplies, so a capable model matters.
And the record does not yet capture the software environment, which is the largest gap in its own
reproducibility story and the next thing we intend to close.
The system is free, open source under the MIT license, and in beta, which means the most useful
contribution a reader can make is to use it and say what broke.
Install it, give it one real analysis, and look at what is left in the folder afterwards.
Skills and specialists are written in plain language, so a procedure your field relies on can be
added by describing it, and issues and contributions are welcome in the repository
\citep{kdensebyok}.

\section*{Availability}

K-Dense BYOK is open source under the MIT license at
\url{https://github.com/K-Dense-AI/k-dense-byok} \citep{kdensebyok}.
The version described in this paper is release 0.9.14, published on 2 September 2026, and the code
carries an automated test suite that runs on every change; installation and model access are
described in Section~\ref{sec:capabilities}.
The agent is built on Pi, an open-source agent toolkit released under the MIT license
\citep{pi2025}, with delegation to specialists and web access supplied by separate community
packages for it.
The scientific skills come from the openly licensed Scientific Agent Skills library
\citep{Kassis2026skills}.
Questions, bug reports, and contributions are welcome through the repository's issue tracker.

\section*{Data and code availability}

The repository above contains only the open-source code of K-Dense BYOK; it does not
contain this study's benchmark data, result bundles, or judge outputs.
The twenty benchmark prompts, the version~3.0 scoring rubric, and the per-prompt scores behind every
figure and table (Table~\ref{tab:perprompt}) are reproduced in full in the appendices to this
manuscript.
The frozen result bundles and the judge's output for each bundle are available from the authors on
request.
The K-Bench~01 benchmark is described in \citet{kbench01}.

\section*{Use of K-Dense BYOK in preparing this manuscript}

This manuscript was prepared with the help of K-Dense BYOK, the system it describes.
The authors used it while searching the literature, working with the evaluation results, and drafting
and revising the text.
Every scientific claim, the design of the evaluation, and the interpretation of the results are the
authors' own, and the authors take responsibility for the content of the paper.
We state the circularity rather than leave it implicit: a paper arguing that the software around a
model shapes research output was itself written with help from the system it argues for.
The evaluation is unaffected, because every result bundle was frozen before scoring, but readers
should weigh this alongside the competing interests declared below.

\section*{Competing interests}

All authors are employees of K-Dense Inc., which develops the K-Dense BYOK platform described and
evaluated in this study.
K-Dense Inc.\ also sells K-Dense Web, a hosted, managed research platform of the kind this paper
argues against, so the authors have a commercial interest in both families of tools.
Independent replication is warranted.

\bibliographystyle{plainnat}
\bibliography{references}

\clearpage
\appendix

\section{Additional evaluation figures}
\label{app:figures}

This appendix gives the prompt-level and criterion-level detail behind the summaries in
Section~\ref{sec:results}, drawn from the same sixty bundles and the same judging pass.
Table~\ref{tab:perprompt} lists the scores of every bundle; the figures that follow display them.

\begin{table}[!htb]
\centering
\caption{Scores of all sixty bundles, as \% of available points for scientific quality (out of 75) and
research execution (out of 25) and as points out of 100 for the secondary combined score, rounded to
whole numbers. Prompts are numbered as in Appendix~\ref{app:prompts}. KD, K-Dense BYOK; CS, Claude
Science; BL, Biomni Lab.}
\label{tab:perprompt}
\small
\begin{tabular}{@{}r rrr rrr rrr@{}}
\toprule
 & \multicolumn{3}{c}{Scientific quality} & \multicolumn{3}{c}{Research execution} & \multicolumn{3}{c}{Combined} \\
\cmidrule(lr){2-4} \cmidrule(lr){5-7} \cmidrule(l){8-10}
Prompt & KD & CS & BL & KD & CS & BL & KD & CS & BL \\
\midrule
1 & 82 & 70 & 68 & 62 & 24 & 50 & 77 & 59 & 64 \\
2 & 88 & 34 & 19 & 62 & 12 & 8 & 82 & 28 & 16 \\
3 & 68 & 74 & 66 & 47 & 17 & 37 & 63 & 60 & 58 \\
4 & 84 & 73 & 61 & 70 & 24 & 37 & 80 & 61 & 55 \\
5 & 82 & 72 & 69 & 70 & 24 & 37 & 79 & 60 & 61 \\
6 & 76 & 57 & 64 & 55 & 29 & 37 & 71 & 50 & 58 \\
7 & 83 & 72 & 59 & 55 & 37 & 37 & 76 & 64 & 54 \\
8 & 88 & 73 & 64 & 63 & 24 & 37 & 82 & 60 & 57 \\
9 & 82 & 73 & 80 & 70 & 29 & 55 & 79 & 62 & 74 \\
10 & 97 & 84 & 68 & 63 & 29 & 42 & 88 & 70 & 62 \\
11 & 85 & 75 & 56 & 70 & 17 & 45 & 81 & 60 & 53 \\
12 & 72 & 75 & 74 & 70 & 17 & 42 & 72 & 60 & 66 \\
13 & 82 & 85 & 74 & 62 & 34 & 42 & 77 & 72 & 66 \\
14 & 79 & 77 & 21 & 45 & 29 & 13 & 70 & 65 & 19 \\
15 & 94 & 67 & 70 & 70 & 24 & 42 & 88 & 56 & 63 \\
16 & 100 & 100 & 90 & 70 & 50 & 42 & 92 & 88 & 78 \\
17 & 98 & 77 & 100 & 70 & 24 & 42 & 91 & 64 & 86 \\
18 & 86 & 79 & 80 & 63 & 24 & 42 & 80 & 65 & 71 \\
19 & 100 & 93 & 40 & 83 & 24 & 37 & 96 & 76 & 39 \\
20 & 94 & 86 & 74 & 70 & 29 & 42 & 88 & 72 & 66 \\
\midrule
Mean & 86.0 & 74.8 & 64.9 & 64.5 & 26.0 & 38.3 & 80.6 & 62.6 & 58.3 \\
\bottomrule
\end{tabular}
\end{table}

\begin{figure}[!htb]
\centering
\includegraphics[width=0.62\textwidth,trim=0 21 0 60,clip]{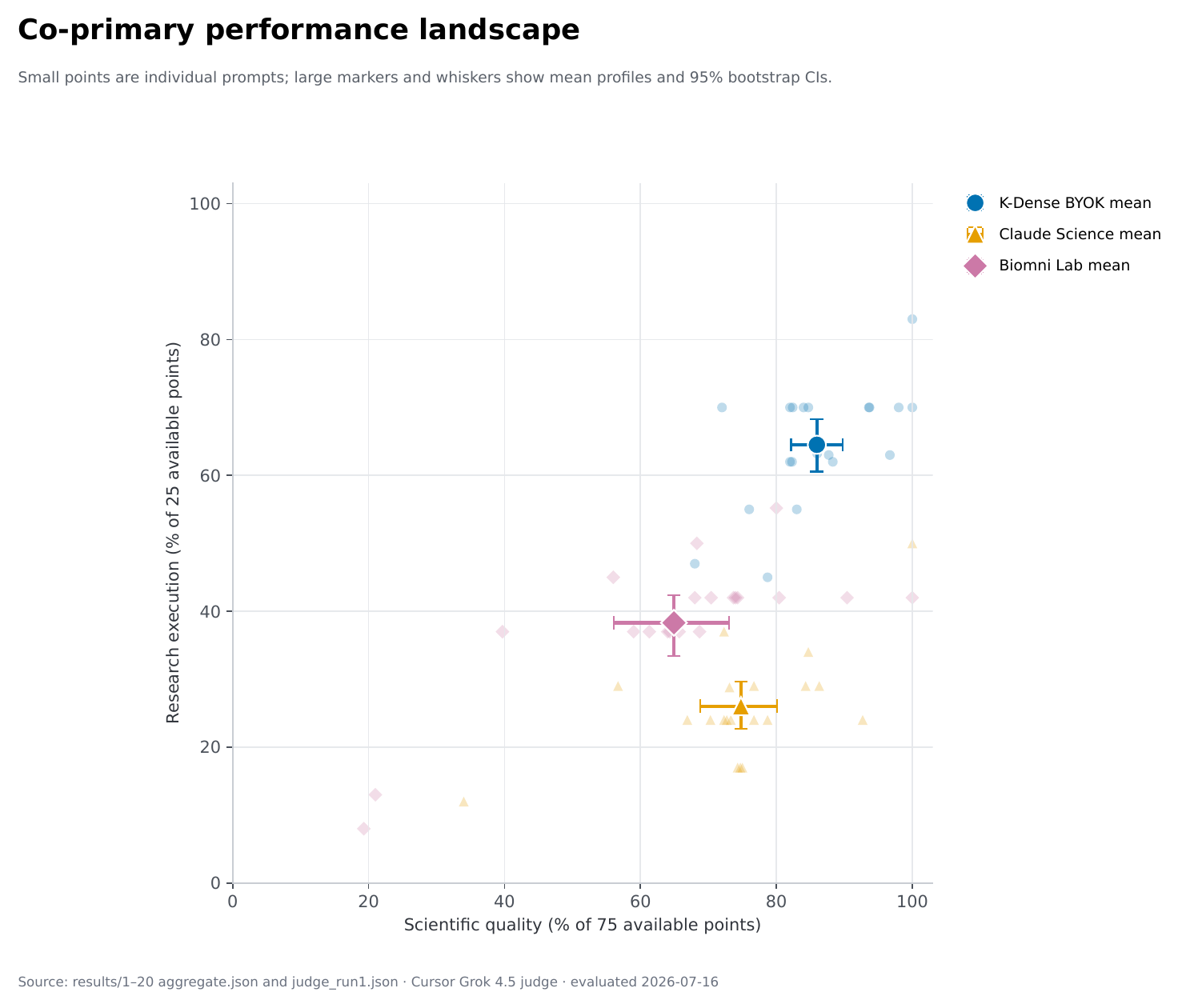}
\caption{The two outcomes plotted against each other, without combining them into one score.
Small points are results on individual prompts; large markers with whiskers are platform means with
95\% bootstrap confidence intervals.
K-Dense BYOK separates from both comparison platforms mainly along the research-execution axis,
while Claude Science and Biomni Lab swap places between the two axes.}
\label{fig:landscape}
\end{figure}

\begin{figure}[!htb]
\centering
\includegraphics[width=\textwidth,trim=0 21 0 54,clip]{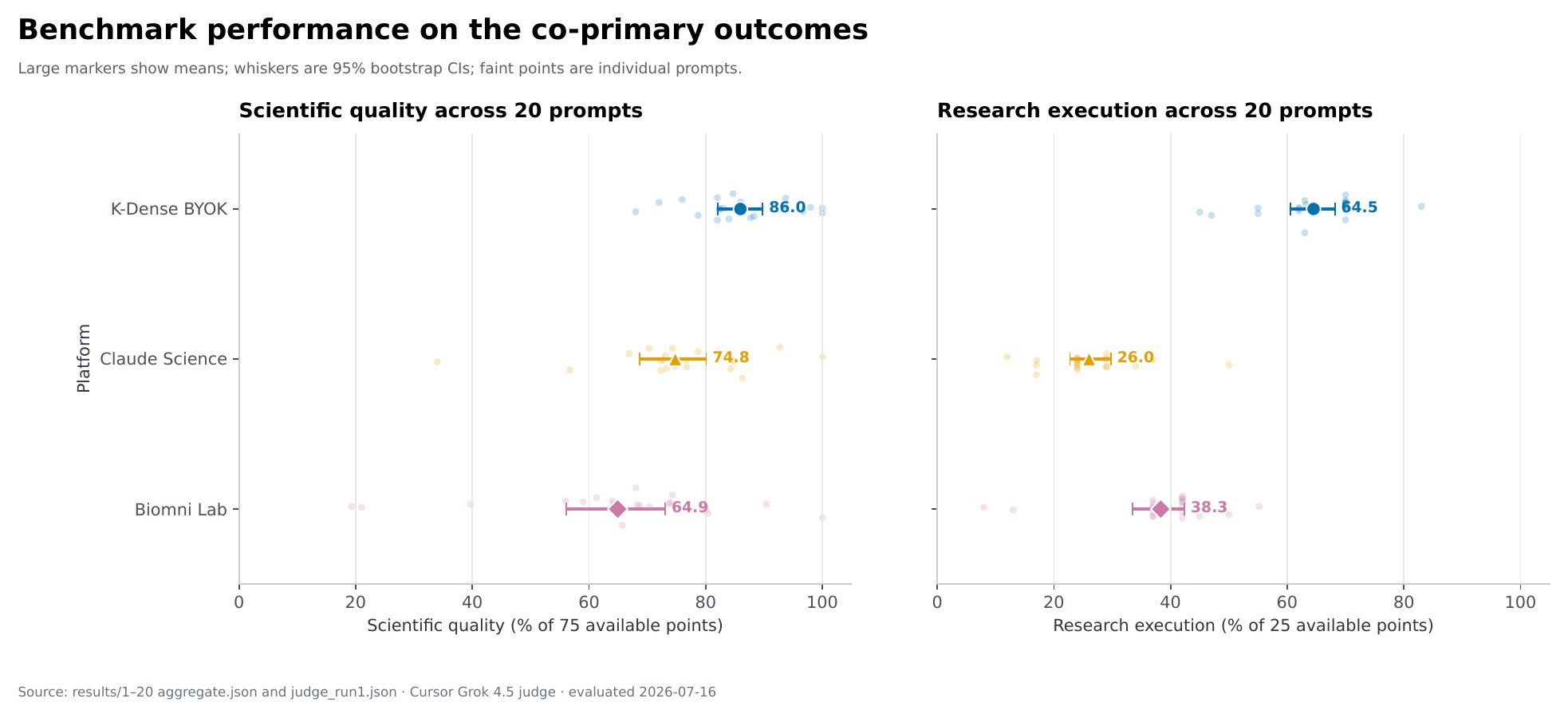}
\caption{Both outcomes across the twenty benchmark prompts.
Large markers show platform means with 95\% bootstrap confidence intervals over prompts; faint
points are the scores on individual prompts.
Left panel: scientific quality as \% of 75 available points.
Right panel: research execution as \% of 25 available points.
The spread between platforms is visibly wider on research execution than on scientific quality.}
\label{fig:overview}
\end{figure}

\begin{figure}[!htb]
\centering
\includegraphics[width=\textwidth,trim=0 21 0 56,clip]{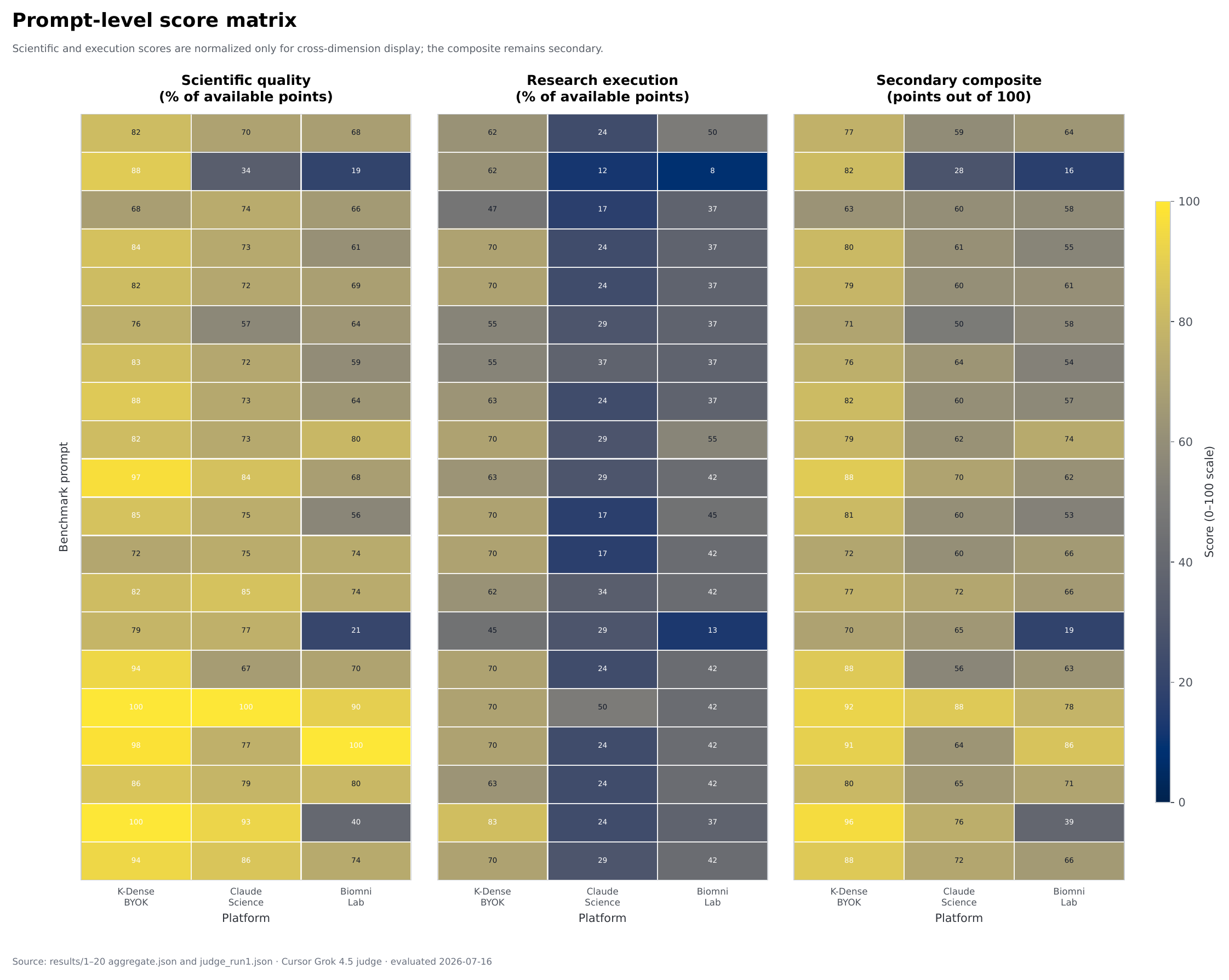}
\caption{Scores for each of the twenty benchmark prompts.
Each row is one prompt and each column one platform; cells give scientific quality and research
execution as percentages of available points, and the secondary combined score in points out of 100,
all on one shared 0--100 color scale that is used for display only.
The research-execution panel separates the platforms on nearly every prompt, while in the
scientific-quality panel the platforms trade places on several prompts.}
\label{fig:heatmaps}
\end{figure}

\begin{figure}[!htb]
\centering
\includegraphics[width=\textwidth,trim=0 21 0 58,clip]{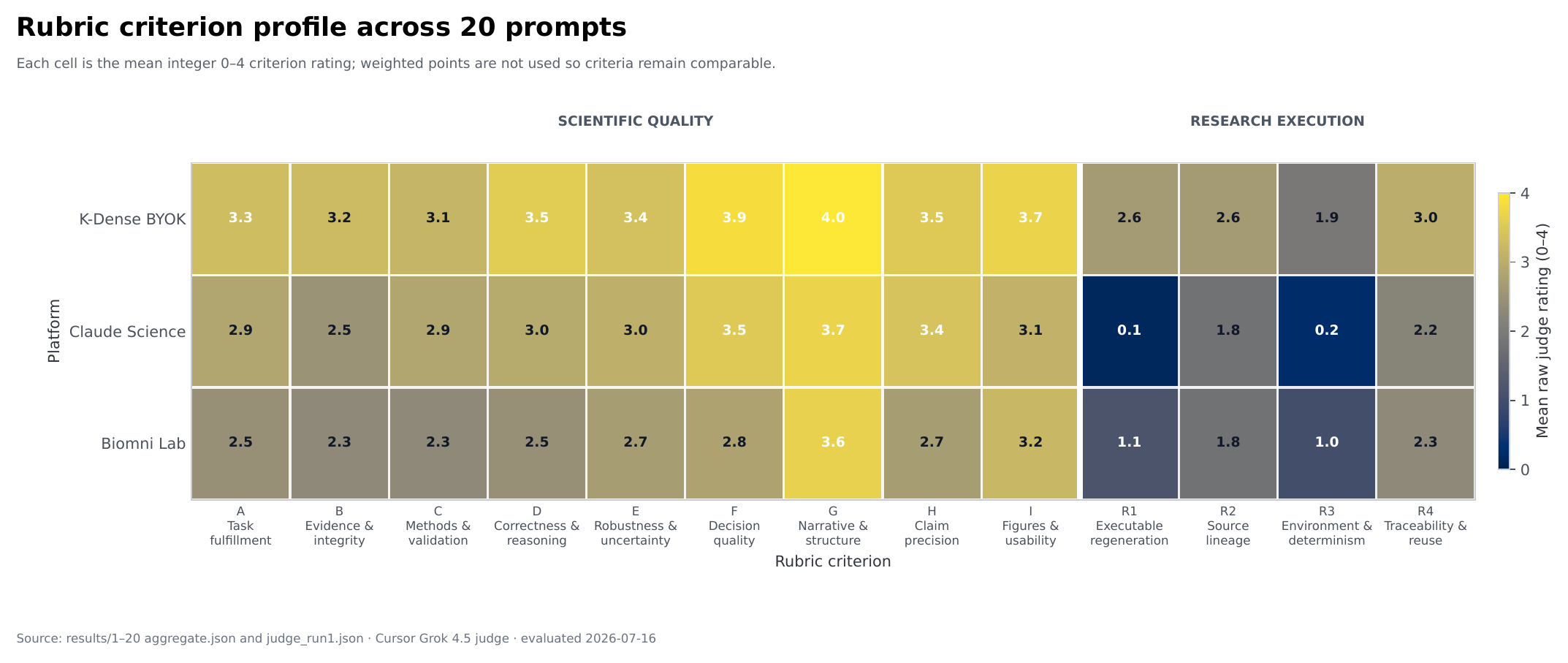}
\caption{Mean rating on each criterion over the twenty prompts.
Cells show mean judge ratings on the 0--4 scale, without weighting, so the criteria can be compared
directly.
Scientific-quality criteria (A--I) are at left and research-execution criteria (R1--R4) at right.
K-Dense BYOK leads on all nine quality criteria but by modest margins.
The main separation is concentrated on executable regeneration (R1) and environment and determinism
(R3).}
\label{fig:criteria}
\end{figure}

\clearpage
\section{Benchmark prompts}
\label{app:prompts}
\sloppy

The twenty prompts below constitute the benchmark task set. Each was presented verbatim to all three platforms. Prompts were authored with GPT-5.6~Sol and are deliberately underspecified in the manner of real scientific requests.

\subsection*{Prompt 1. Oncology and Medicinal Chemistry}
Using the open masked-mutation and RNA-expression data for GDC project TCGA-LUAD from Data Release 45.0, ChEMBL 37 assays for EGFR (\texttt{CHEMBL203}), and the osimertinib-bound EGFR structure PDB \texttt{4ZAU}, identify one inhibitor chemotype and a 4--6-mutation biochemical panel for a preclinical resistance study. Separate biochemical from cellular assays, normalize comparable potency measurements, cluster compounds by scaffold, and integrate assay quality with tumor variant prevalence and expression. Deliver a ranked chemotype--mutation matrix, the nominated series and mutation panel, a structural rationale, and explicit go/no-go uncertainties; do not make patient-treatment recommendations.

\subsection*{Prompt 2. Infectious-Disease Genomics and Structural Biology}
Using the corrected WHO 2023 \emph{Mycobacterium tuberculosis} mutation catalogue (ISBN \texttt{9789240082410}, files pinned at commit \texttt{0bb3914348c5a4c981859601447834c08f03ee3d}) and the rifampicin-bound RNA-polymerase structure PDB \texttt{5UHB}, select 8--12 \emph{rpoB} substitutions for a surveillance and laboratory-phenotyping panel. Combine WHO confidence grades, isolate counts, and association uncertainty with residue alignment, ligand distance, contact disruption, and local structural environment so that the panel covers both common resistance alleles and mechanistically distinct unresolved variants. Deliver the prioritized panel, an epidemiological evidence table, a structure map, and phenotyping controls; do not recommend therapy.

\subsection*{Prompt 3. Immunology and Biomaterials}
Using GEO \texttt{GSE248524} and the associated source data in PMCID \texttt{PMC11315930}, determine whether the next wound-scaffold experiment should advance the lightly crosslinked hydrogel, retain the heavily crosslinked formulation, or test an intermediate. Perform donor-aware pseudobulk and differential-abundance analyses, distinguish macrophage and fibroblast state changes from compositional shifts, and integrate those results with measured crosslinking, mechanics, degradation, uptake, and tissue infiltration. Nominate one immune--stromal mechanism for perturbation only if it is robust to sensitivity analyses. Deliver a formulation decision, supporting cell-state and material-property figures, one mechanistic experiment, and a concise statement of causal limitations.

\subsection*{Prompt 4. Pharmacogenomics and Population Genetics}
Using the GRCh37 1000 Genomes Phase 3 chromosome-10 release dated \texttt{20130502} and its sample panel, dbSNP Build 157 records \texttt{rs4244285}, \texttt{rs4986893}, and \texttt{rs12248560}, and ClinPGx/CPIC guideline record \texttt{PA166251443}, decide whether a low-cost research-only CYP2C19 assay should use a universal \texttt{*2/*3/*17} core or population-specific add-ons. Estimate phased allele and haplotype frequencies by population, audit missingness and Hardy--Weinberg deviations, test whether each SNP tags its stated star allele, and compare total versus worst-population coverage. Deliver the assay specification, per-population coverage with uncertainty, an equity audit, and the evidence that would change the design; do not provide prescribing advice.

\subsection*{Prompt 5. Developmental Toxicology and Placental Biology}
Using EPA ToxCast \texttt{invitroDB v4.2} Version 13 for PFOS (\texttt{DTXSID3031864}) and PFOA (\texttt{DTXSID8031865}) together with the first-trimester maternal--fetal-interface atlas \texttt{E-MTAB-6701}, choose one compound, one trophoblast or decidual cell state, and one molecular endpoint for a mechanistic placental-model experiment. Filter ToxCast curves for assay quality, cytotoxicity, and nonspecific activity, then map credible targets to donor-aware expression and developmental specificity across 6--12 gestational weeks. Deliver a ranked chemical--cell-state--pathway table, the nominated organoid or trophoblast experiment with controls, and separate evidence and speculation sections; do not infer individual risk.

\subsection*{Prompt 6. Single-Cell Biology and Translational Medicine}
Using ileal Crohn's scRNA-seq \texttt{GSE134809} and the independent pediatric inception cohort \texttt{GSE134881}/\texttt{SRP216403}, derive a cell-state-informed signature of anti-TNF nonresponse and decide whether it merits prospective assay validation. Perform patient-level QC, annotation, pseudobulk contrasts, and signature derivation only in the single-cell cohort; lock the gene panel before testing it against the public response labels in the bulk cohort. Quantify cohort shift, calibration, discrimination, and uncertainty while preventing patient leakage and adjusting for baseline disease activity. Deliver the locked panel, an analysis audit, validation plots, and an advance/redesign/stop recommendation; do not make individual treatment recommendations.

\subsection*{Prompt 7. Radiology and Genomics}
Using TCIA's \texttt{NSCLC-Radiogenomics} collection Version 4 (\texttt{10.7937/K9/TCIA.2017.7hs46erv}) and matched RNA-seq \texttt{GSE103584}/\texttt{SRP117020}, test whether a compact CT-radiomics panel adds reproducible information about tumor pathway activity beyond radiologist semantic annotations. Include only subjects with matched preoperative CT, valid tumor segmentation, and RNA data; standardize image handling, extract segmentation-robust features, derive pathway scores within training folds, and compare semantic-only, radiomics-only, and combined models with nested patient-level validation. Deliver a cohort attrition diagram, locked feature specification, out-of-fold performance with uncertainty, and an advance/retain-semantics/stop decision for external validation.

\subsection*{Prompt 8. Proteomics and Cardiology}
Using ProteomeXchange/PRIDE \texttt{PXD008934}, revision 3, identify left-ventricular protein modules that distinguish decompensated heart failure from normal or compensated hypertrophy and remain directionally consistent across ischemic, dilated, and hypertrophic etiologies. Reproduce contaminant and missingness filters on the processed LFQ data, fit age- and sex-adjusted empirical-Bayes contrasts, quantify cross-etiology heterogeneity, and evaluate module stability by leave-one-heart-out analysis and bootstrap resampling. Deliver cross-etiology effect plots, a short ranked module and protein list, and a validate/replicate-first/deprioritize recommendation with an orthogonal experimental plan; do not infer post-translational modifications from abundance data.

\subsection*{Prompt 9. Microbiome Science and Nutrition}
Using the fixed 2018 American Gut fecal sOTU table (\texttt{10.6084/m9.figshare.6137192}) and mapping file (\texttt{10.6084/m9.figshare.6137315}), determine whether consuming more than 30 versus 10 or fewer plant types per week is robustly associated with gut microbial ecology after adjustment for antibiotics, age, BMI, geography, sequencing plate, and other dietary variables. Analyze one stool sample per participant with compositional or phylogenetic balances, constrained-permutation beta-diversity tests, missingness analysis, and leave-one-country-out validation. Deliver adjusted effects, balance and transportability diagnostics, and a fund/redesign/do-not-fund decision for a controlled feeding study, including one prespecified microbial endpoint; do not infer microbial function from 16S data alone.

\subsection*{Prompt 10. Digital Pathology and Spatial Statistics}
Using Version 2 of the breast-cancer imaging-mass-cytometry dataset \texttt{10.5281/zenodo.4607374}, determine whether tumor, stromal, and immune-cell organization adds stable prognostic information beyond cell composition and standard pathology variables. Validate phenotypes, masks, and coordinates on a stratified image subset; compute edge-corrected marked spatial statistics and neighborhood enrichment with within-image label permutations; aggregate repeated images at the patient level; and compare composition-only with composition-plus-spatial survival models using patient-grouped nested validation. Deliver QC overlays, a locked spatial-feature panel, incremental out-of-fold performance with uncertainty, and an advance/simplify/stop decision for external validation rather than a clinical predictor.

\subsection*{Prompt 11. Plant Genomics and Climate Adaptation}
Using the \emph{Arabidopsis thaliana} 1001 Genomes 1,135-accession release GMI-MPI v3.1 and WorldClim v2.1 1970--2000 normals, nominate 24 accessions for a drought-by-heat common-garden experiment. Restrict variant extraction to a preregistered stress and phenology gene panel, control genotype--climate associations for ancestry, meta-analyze across ancestry groups, and use BIO5, BIO6, BIO14, and BIO17 at accession coordinates. Select accessions by maximin diversity across genotype, provenance climate, and ancestry rather than association strength alone. Deliver the accession panel, locus-level hypotheses with uncertainty, population-structure diagnostics, and a factorial drought $\times$ heat experiment with matched controls.

\subsection*{Prompt 12. Remote Sensing and Crop Science}
Within Iowa ROI \texttt{[-93.75, 41.85, -93.15, 42.22]}, use Sentinel-2 Collection 1 L2A tile \texttt{T15TVG} acquisitions from 2023-04-06, 05-24, 06-20, 07-10, 08-22, 09-08, and 10-21 together with the 2023 USDA Cropland Data Layer to rank twenty 1-km corn or soybean cells for field scouting. Apply cloud and shadow masks, retain crop-pure pixels, derive NDVI/NDRE establishment, peak-vigor, persistence, and senescence metrics, and pair each robust crop-specific anomaly with a nearby normal control. Deliver a geospatial ranked list, phenology plots, uncertainty flags, and a nitrogen $\times$ water follow-up design; do not claim a causal stress diagnosis from imagery alone.

\subsection*{Prompt 13. Marine Ecology and Oceanography}
Using the Global Coral Bleaching Database NCEI accession \texttt{0228498} Version 1.1 and NOAA daily OISST v2.1 for 1982--2021, identify repeatedly surveyed reefs that bleached less or more than expected from antecedent heat exposure. For each survey, calculate 84-day cumulative positive thermal anomaly, maximum anomaly, and event duration relative to a fixed 1982--2011 local seasonal baseline; fit a grouped model with region and year effects and validate by site. Rank only repeat-survey locations. Deliver resilient and sensitive candidate lists, observed-versus-expected plots with uncertainty, mechanistic hypotheses, and a plan for temperature loggers, symbiont profiling, and controlled heat-tolerance assays.

\subsection*{Prompt 14. Conservation Biology and Infectious-Disease Ecology}
Using the USGS national Bd/Bsal survey Version 2.0 (\texttt{10.5066/P9BGQA1T}), USGS GAP Species Range Maps CONUS 2001 v1 (\texttt{10.5066/F7Q81B3R}), and the USGS Watershed Boundary Dataset, prioritize 25 HUC12 watersheds and focal salamander species for a prospective Bsal early-detection survey. Reconcile taxonomy, quantify sampling intensity and unsampled range, weight narrow-ranging taxa, and use set-cover optimization to maximize complementary species coverage. Allocate enough swabs per site for 95\% detection probability at 5\% prevalence, adjusted for assay sensitivity. Deliver the ranked sites and species, total sampling allocation, detection assumptions, and field/QC protocol; treat GAP ranges as design priors rather than current occupancy and do not present a Bsal occurrence-risk map.

\subsection*{Prompt 15. Microbial Ecology and Environmental Chemistry}
Using Tara Oceans MAG distributions from Figshare \texttt{10.6084/m9.figshare.4902938.v3}, nutrient measurements \texttt{10.1594/PANGAEA.875575}, and sample registry \texttt{10.1594/PANGAEA.875582}, identify ten MAGs whose distributions show stable associations with nitrate, phosphate, silicate, or nutrient stoichiometry across ocean regions. Audit sample joins, filter rare MAGs, transform compositional abundances, fit nutrient-association models with false-discovery control, and require stability under leave-one-region-out analysis. Connect robust hits to available functional annotations without treating association as causation. Deliver the ranked MAGs and environments, effect-stability evidence, alternative explanations, and a nitrate $\times$ phosphate microcosm experiment with a qPCR or metagenomic readout.

\subsection*{Prompt 16. Climatology and Epidemiology}
Using Daymet V4 R1 daily temperature for 1991--2020 (\texttt{Daymet\_Daily\_V4R1\_2129}) and CDC PLACES County Data 2024 (\texttt{d3i6-k6z5}), recommend exactly 25 contiguous-U.S. counties for heat-resilience research grants. Define heat burden as the mean national percentile of annual days with maximum temperature at least 35$^\circ$C and nights with minimum temperature at least 20$^\circ$C; define health burden as the mean percentile of age-adjusted CHD, stroke, COPD, diabetes, and disability prevalence; and rank the product of heat burden, health burden, and adult population. Propagate interannual climate variation and PLACES confidence intervals through simulation. Deliver the county table, maps, rank-stability intervals, final allocation, and comparison with heat-only and health-only rankings.

\subsection*{Prompt 17. Air Quality and Labor Economics}
Using EPA AQS daily PM2.5 file \texttt{daily\_88101\_2023.zip} and BLS QCEW \texttt{2023\_annual\_singlefile.zip}, estimate county-level construction worker-days exposed to ambient PM2.5 above 35 $\mu$g/m$^3$ and recommend 20 counties for a worker-protection study. Deduplicate pollutant-standard rows, define each county-day as the median valid daily mean across distinct monitors, join county FIPS to private-sector QCEW construction industry \texttt{1012}, and calculate exposed worker-days and wage-bill exposure while handling monitoring gaps and suppressed employment cells. Deliver county rankings, a concentration curve, sensitivity to median versus maximum monitor exposure, and the selected counties; distinguish potential wage exposure from realized economic loss.

\subsection*{Prompt 18. Hydrology and Infectious-Disease Epidemiology}
Using California's West Nile Virus Cases dataset for 2006--2023 (dataset UUID \texttt{3205b420-3f62-4a02-8d2e-9a9ed34c49f4}) and USGS daily mean discharge parameter \texttt{00060}/statistic \texttt{00003}, test whether antecedent streamflow improves prediction of July--December county outbreaks, defined as at least five cases, beyond a non-hydrologic baseline. Retain gauges with at least 80\% coverage, standardize flow within gauge and day of year, construct low-flow and dry-to-wet-pulse predictors, and use county/year effects with leave-one-year-out validation. Advance a hydrology trigger only if AUPRC improves by at least 0.05 and calibration slope is 0.8--1.2. Deliver the gauge audit, model comparison, decision, and ten prioritized counties with limitations.

\subsection*{Prompt 19. Seismology and Infrastructure Engineering}
Reconstruct bridge-inspection priorities after the 24 August 2014 South Napa earthquake using USGS event \texttt{nc72282711}, reviewed ShakeMap Atlas product \texttt{nc72282711/atlas/1624995941193}, the 2014 California National Bridge Inventory, and FEMA Hazus 6.1 bridge fragilities. Interpolate PGA, SA(0.3 s), and SA(1.0 s) at bridges within MMI VI or greater, map inventory attributes to Hazus bridge classes, propagate shaking and fragility uncertainty, and calculate expected loss-of-function days weighted by average daily traffic. Deliver the exact 20 NBI structure IDs to inspect first, their expected damage and downtime, and the share of predicted network vehicle-days captured, while stating omitted ground-failure and rerouting effects.

\subsection*{Prompt 20. Urban Heat, Demography, and Energy Equity}
Allocate 1,000 cooling and weatherization research slots across 2022 census tracts intersecting Richmond, Virginia, using the 2021 NIHHIS--CAPA Richmond heat rasters (OSF project \texttt{3xvmg}, raster file \texttt{kw5bc}) and DOE LEAD 2022 Virginia data (\texttt{10.25984/2504170}). Summarize afternoon temperature, heat index, and evening temperature by tract; combine these with 0--80\% AMI household counts and modeled energy burden; propagate zonal and LEAD uncertainty; and compare joint-priority, heat-only, and energy-only allocations. Allocate proportionally among selected tracts with at least 25 and at most 200 slots per tract. Deliver tract GEOIDs, final counts, three maps, and sensitivity results; treat the heat raster as a campaign snapshot and not household-level risk.

\clearpage
\section{Scoring rubric}
\label{app:rubric}

This appendix reproduces the substance of rubric version~3.0, the frozen instrument used to score every prompt~$\times$ platform result bundle. Long normative passages are condensed, but no weight, scale, or scoring rule is altered.

\subsection{Structure and score construction}

The unit of evaluation is one prompt $\times$ platform result bundle. The rubric has two co-primary outcomes: \emph{scientific quality} (0--75 points, criteria A--I) and \emph{research execution} (0--25 points, criteria R1--R4). Their sum is reported as a secondary 100-point composite. For each criterion the judge assigns an integer score from 0 to 4 and converts it to weighted points:

\begin{equation*}
\text{criterion\_points} = \text{criterion\_weight} \times \frac{\text{raw\_score}}{4},
\end{equation*}

\begin{equation*}
\text{scientific\_quality} = \sum_{A}^{I} \text{points}, \qquad
\text{research\_execution} = \sum_{R1}^{R4} \text{points}.
\end{equation*}

There are no score caps, bonus points, or deductions outside the weighted criteria. The two co-primary outcomes are reported separately and prominently alongside the composite; a platform is not described as generally superior from the composite alone when the two dimensions materially disagree.

\subsection{Criterion weights}

\begin{table}[h]
\centering
\small
\begin{tabular}{@{}llr@{}}
\toprule
ID & Criterion & Points \\
\midrule
A & Task fulfillment & 13 \\
B & Evidence and data integrity & 11 \\
C & Methods and validation & 13 \\
D & Result correctness and reasoning & 11 \\
E & Robustness and uncertainty & 6 \\
F & Decision quality and follow-up & 6 \\
G & Scientific narrative and structure & 5 \\
H & Claim precision and epistemic honesty & 5 \\
I & Figures, tables, and deliverable usability & 5 \\
\midrule
\multicolumn{2}{l}{\textbf{Scientific quality}} & \textbf{75} \\
\midrule
R1 & Executable regeneration & 8 \\
R2 & Source lineage and acquisition & 7 \\
R3 & Environment and determinism & 5 \\
R4 & Traceability and reusable outputs & 5 \\
\midrule
\multicolumn{2}{l}{\textbf{Research execution}} & \textbf{25} \\
\midrule
\multicolumn{2}{l}{\textbf{Total}} & \textbf{100} \\
\bottomrule
\end{tabular}
\end{table}

The weights are prespecified normative priorities, not empirically estimated constants. Task fulfillment and methods receive the largest scientific weights because failure in either can invalidate the requested study; within research execution, regeneration and source lineage receive the greatest weight because they most directly test whether a claimed result can be recovered and connected to its inputs.

\subsection{Universal 0--4 anchors}

\begin{table}[h]
\centering
\small
\begin{tabular}{@{}rp{12.2cm}@{}}
\toprule
Score & Anchor \\
\midrule
4 & Complete, correct, specific, and auditable. All central requirements are met and no material defect is found. \\
3 & Substantially correct and complete. Only minor omissions or defects that do not change the conclusion. \\
2 & Partially correct. A material omission, weak choice, or error reduces confidence, but useful work remains. \\
1 & Minimal or seriously flawed. Major requirements are absent, unsupported, or incorrect. \\
0 & Absent, nonresponsive, contradicted by evidence, fabricated, or fundamentally wrong. \\
\bottomrule
\end{tabular}
\end{table}

Integer scores only. A score of 4 requires affirmative evidence, not merely the absence of a detected problem. For the communication criteria (G--I), the anchors are interpreted as quality of scientific communication rather than analytical correctness: a polished but overclaiming report is not a 4, whereas a concise, precise report with clear caveats can be.

\subsection{Evidence boundary}

The judge receives the exact prompt, the frozen result bundle, a manifest of included files, a pointer to the main report, and the same tool and source access for every platform. Scores are based only on evidence present in the frozen bundle or directly verifiable from its cited sources. The judge does not assume an analysis occurred because a report describes it; missing evidence receives no assumed credit. Before scoring, the prompt text, rubric, per-prompt requirement checklist, every bundle and manifest, the judge model and settings, and the platform evaluation order are frozen.

For each prompt the judge extracts and freezes a requirement checklist in five categories---inputs and scope; analytical work; comparisons and validation; deliverables and decision; and constraints---marking each requirement \texttt{CENTRAL} or \texttt{SUPPORTING}. One checklist is used for all platforms on the same prompt, and requirement extraction is not itself a scored platform task.

\subsection{Standard result-bundle contract}

Before evaluation, every platform submits a functionally equivalent bundle. Exact filenames may differ, but each bundle must provide a README naming the main report and a documented regeneration command; a manifest listing included files with cryptographic hashes; an executable entry point for the decision-critical workflow; an environment specification such as a lockfile or container definition; machine-readable decision-critical results with stable identifiers and units; source receipts and transformation records for decision-critical inputs; and execution logs disclosing failures, retries, substitutions, and manual interventions. Absence of an item is scored under the relevant criterion rather than triggering an automatic score cap. A preflight validator inventories these items before judging but must not infer scientific correctness from their presence.

\subsection{Scientific-quality criteria (A--I)}

\textbf{A. Task fulfillment (13).} Whether required inputs and scope are used, required analyses performed, required comparisons and validation completed, requested deliverables and decision supplied, and explicit constraints followed. Missing a central requirement normally prevents a score above 2; describing an analysis without executing it does not satisfy a request to perform it.

\textbf{B. Evidence and data integrity (11).} Whether required sources, versions, identifiers, and cohorts are correct; joins, filters, units, transformations, exclusions, and denominators are sound; decision-critical claims are supported; and the report agrees with code, tables, figures, and structured results. Citations alone do not establish support.

\textbf{C. Methods and validation (13).} Whether the unit of analysis and study design are correct, methods fit the question and data type, assumptions and confounding and multiplicity and calibration are handled, preprocessing and validation prevent leakage, controls and baselines and holdouts are appropriate, and QC matches the stated method. Proposed-but-unexecuted methods earn no execution credit unless the prompt asks for a plan.

\textbf{D. Result correctness and reasoning (11).} Whether reported values are computationally and statistically consistent, the conclusion follows from the measured effects and uncertainty, association and prediction and mechanism and causation are distinguished, claims remain within the evidence, and surprising findings receive checks. Scientific plausibility is not a substitute for verification.

\textbf{E. Robustness and uncertainty (6).} Whether the work quantifies uncertainty for decision-critical results, performs sensitivity or stability checks, examines subgroup or transportability concerns, and identifies material limitations and alternative explanations. Generic limitation boilerplate receives little credit.

\textbf{F. Decision quality and follow-up (6).} Whether the requested decision is explicit and evidence-based, decision rules or thresholds are stated, the follow-up study addresses the main uncertainty, controls and comparators and readouts are appropriate, and the proposed work is feasible and discriminating.

\textbf{G. Scientific narrative and structure (5).} Whether a knowledgeable reader can follow the science without reverse-engineering the bundle: the objective stated early, a coherent path from question to decision, the main result and caveats easy to locate, and an abstract that reflects the body rather than overselling it.

\textbf{H. Claim precision and epistemic honesty (5).} Whether language matches evidence: observations and inferences and speculation separated; association, prediction, mechanism, and causation worded correctly; uncertainty placed near the claims it qualifies; hedging proportionate; and conflicting or fragile findings acknowledged. Confident tone alone is not a defect; overclaiming beyond the evidence is.

\textbf{I. Figures, tables, and deliverable usability (5).} Whether required visual and structured outputs are present and decision-relevant, axes and units and sample sizes and error representations are labeled, captions state what is shown, figures are readable without unmarked companion files, and narrative claims match displayed values. Polish and decorative graphics receive no independent credit.

\subsection{Research-execution criteria (R1--R4)}

Research execution measures whether the work can be inspected, regenerated, and reused; it does not establish that the scientific conclusion is correct. Artifact count and workflow complexity receive no credit by themselves.

\textbf{R1. Executable regeneration (8).} The judge attempts regeneration in a standardized clean environment, running the documented command without modifying scientific logic, and records command, exit status, runtime, interventions, expected and observed values, tolerances, and check results. A~4 requires one documented entry point to regenerate the main decision and all decision-critical outputs within prespecified tolerances without evaluator repair; 3 allows a minor non-scientific intervention such as a path remap; 2 requires substantial setup or partial regeneration with the central result still recoverable; 1 means executable fragments exist but the main decision cannot be regenerated; 0 means no substantive executable workflow exists. Scientific logic is never silently repaired to award execution credit.

\textbf{R2. Source lineage and acquisition (7).} Whether the bundle records what was actually obtained and used---source locations, versions, retrieval dates, stable identifiers, licenses, cryptographic hashes, acquisition commands, transformations, exclusions, and substitutions. A~4 requires a verifiable receipt, version, retrieval record, hash, and complete transformation lineage for every decision-critical input; 0 means decision-critical provenance is absent, contradicted, or fabricated. Transparent documentation of a wrong source earns execution credit here but still loses scientific credit under A and B.

\textbf{R3. Environment and determinism (5).} Whether dependencies, runtime information, parameters, seeds, deterministic settings, configuration, path assumptions, and resource requirements suffice to reproduce the recorded execution. A~4 requires a lockfile or container plus runtime specification, parameters, seeds, and resource requirements supporting a one-command bootstrap; 0 means no usable environment information is supplied.

\textbf{R4. Traceability and reusable outputs (5).} Whether each central requirement and decision-critical claim can be traced backward through report, machine-readable result, code, transformation, and source record; whether failures, fallbacks, substitutions, retries, and manual interventions are disclosed near the affected outputs; and whether decision-relevant outputs use stable identifiers, units, labels, and schemas. A~4 requires a complete requirement~$\rightarrow$~claim~$\rightarrow$~result~$\rightarrow$~code~$\rightarrow$~source chain with internally consistent structured outputs and failure records.

\subsection{Judging protocol}

The v3.0 protocol uses one fresh-context judge run per prompt $\times$ platform bundle, with weighted scores calculated directly from that run's raw criterion scores. Platforms are evaluated one at a time under the same prompt checklist, judge settings, evidence rules, and access policy, using the same clean environment, network policy, resource limits, timeout policy, and permitted evaluator interventions. Deviations are recorded as evaluator access failures rather than silently repairing one platform more than another. All prompts receive equal weight in platform-level summaries, and platform comparisons use paired prompt-level results.

\end{document}